%% file: main.tex
\documentclass[runningheads]{llncs}
\usepackage{graphicx}
\usepackage{comment}
\usepackage{amsmath,amssymb} 
\usepackage{color}

\input{setup/package}
\input{setup/macros}

\input{setup/symbols}

\input{setup/graphicspath}

\begin{document}
\pagestyle{headings}
\mainmatter
\def\ECCVSubNumber{1653}  

\title{Semantic View Synthesis} 

\titlerunning{Semantic View Synthesis}
%
\author{Hsin-Ping Huang$^1$, Hung-Yu Tseng$^2$, Hsin-Ying Lee$^2$, Jia-Bin Huang$^3$}
\authorrunning{H.-P. Huang et al.}
%
\institute{$^1$UT Austin~~~$^2$University of California, Merced~~~$^3$Virginia Tech}
\maketitle
\input{figure/fig_teaser}

\input{0_abstract}
\input{1_introduction}

\input{2_prior}

\input{3_method}

\input{4_result}
\input{5_conclusion}

\clearpage
%
%
\bibliographystyle{splncs04}
\bibliography{references}
\end{document}


\pagestyle{headings}
\mainmatter
\def\ECCVSubNumber{1653}  

\title{Semantic View Synthesis \\
Supplementary Material} 

\titlerunning{ECCV-20 submission ID \ECCVSubNumber} 
\authorrunning{ECCV-20 submission ID \ECCVSubNumber} 
\author{Anonymous ECCV submission}
\institute{Paper ID \ECCVSubNumber}

\maketitle

\input{rebut_content}

\clearpage
%
%
\bibliographystyle{splncs04}
\bibliography{references}

%% file: setup/package.tex
\usepackage{graphicx}
\usepackage{subcaption}
\usepackage{float}
\usepackage{lscape}                                         
\usepackage{wrapfig}

\usepackage[lined,ruled,linesnumbered]{algorithm2e}
\usepackage{animate} 

\usepackage{booktabs}                   
\usepackage{multirow}

\usepackage{makecell}

\usepackage{paralist}
\usepackage{enumitem}

\usepackage{bm}                          
\usepackage{epsfig}                      
\usepackage{graphicx}                  
\usepackage{mathtools}
\usepackage{amssymb,amsmath}   

\usepackage{units}
\usepackage{color}

\usepackage{comment}

\usepackage{url}  
\usepackage[pagebackref=true,breaklinks=true,letterpaper=true,colorlinks,bookmarks=false]{hyperref}

\usepackage{xspace}
\usepackage[table]{xcolor}
\usepackage{setspace}


%% file: setup/macros.tex
\def\etal{et~al.}			  
\def\eg{e.g.,~}               
\def\ie{i.e.,~}               

\newlength\paramargin
\newlength\figmargin

\newlength\secmargin
\newlength\figcapmargin
\newlength\tabcapmargin

\newcommand{\mpage}[2]
{
\begin{minipage}{#1\linewidth}\centering
#2
\end{minipage}
}

\newcommand{\topic}[1]
{
\vspace{1mm}\noindent\textbf{#1}
}

\newcommand{\secref}[1]{Section~\ref{sec:#1}}
\newcommand{\figref}[1]{Figure~\ref{fig:#1}} 
\newcommand{\tabref}[1]{Table~\ref{tab:#1}}

\newcommand{\subsecref}[1]{Section~\ref{subsec:#1}}

\long\def\ignorethis#1{}

\newcommand {\james}[1]{{\color{magenta}\textbf{James: }#1}\normalfont}

\newcommand{\tb}[1]{\textbf{#1}}

{}

\newbox\jsavebox%

%% file: setup/symbols.tex
\def\xi{\mathbf{x}_i}

%% file: setup/graphicspath.tex
\graphicspath{{figure}, {example}}

%% file: figure/fig_teaser.tex
\newlength\fta
\setlength\fta{0.155\linewidth}
\begin{figure}[th]
\centering
\includegraphics[width=\fta, height=\fta]{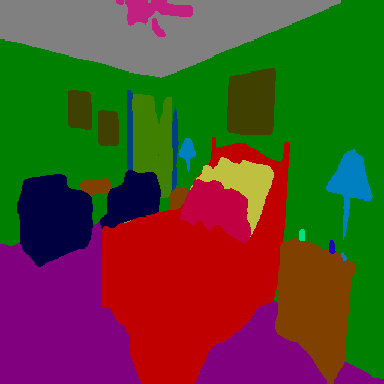}
\includegraphics[width=\fta, height=\fta]{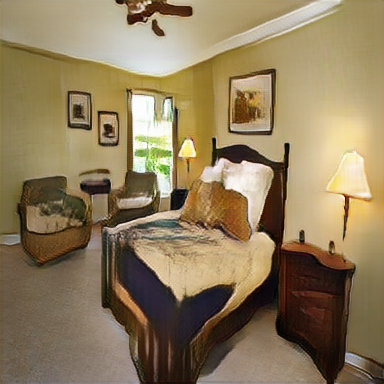}
\includegraphics[width=\fta, height=\fta]{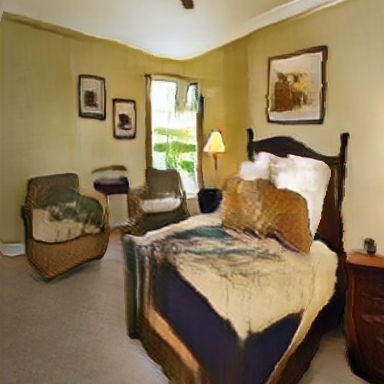}\hfill
\includegraphics[width=\fta, height=\fta]{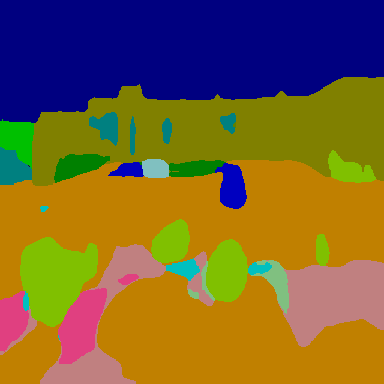}
\includegraphics[width=\fta, height=\fta]{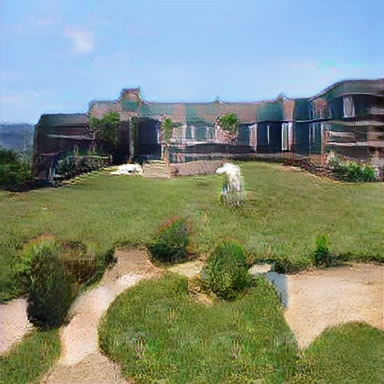}
\includegraphics[width=\fta, height=\fta]{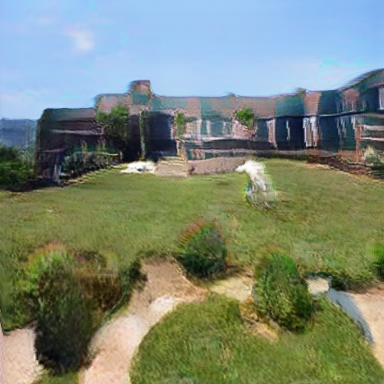}

\includegraphics[width=\fta, height=\fta]{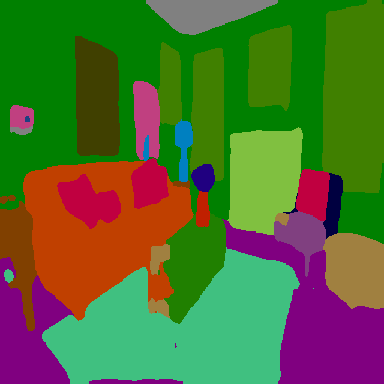}
\includegraphics[width=\fta, height=\fta]{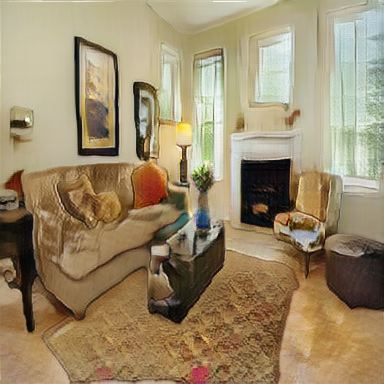}
\includegraphics[width=\fta, height=\fta]{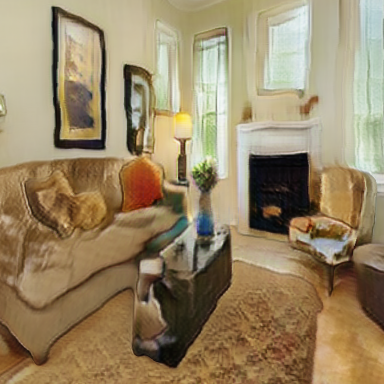}\hfill
\includegraphics[width=\fta, height=\fta]{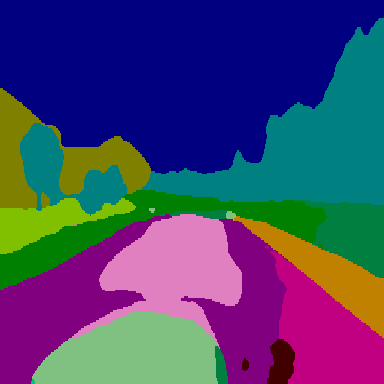}
\includegraphics[width=\fta, height=\fta]{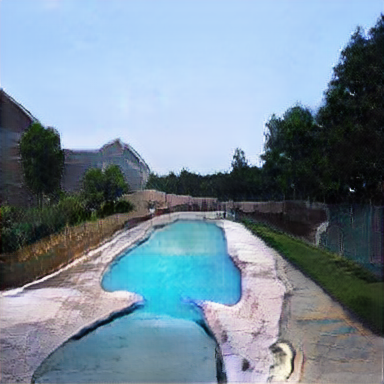}
\includegraphics[width=\fta, height=\fta]{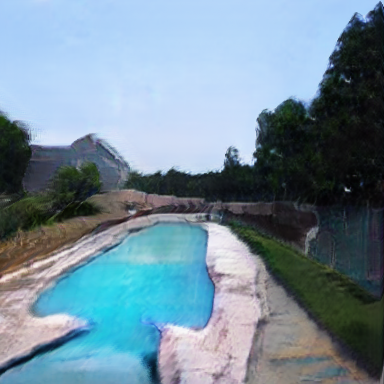}

\includegraphics[width=\fta, height=\fta]{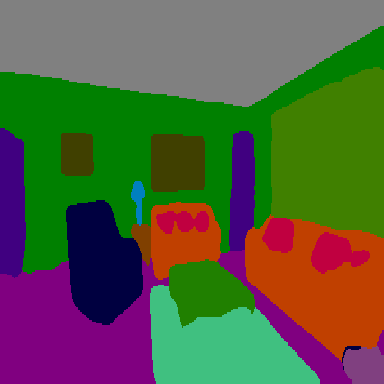}
\includegraphics[width=\fta, height=\fta]{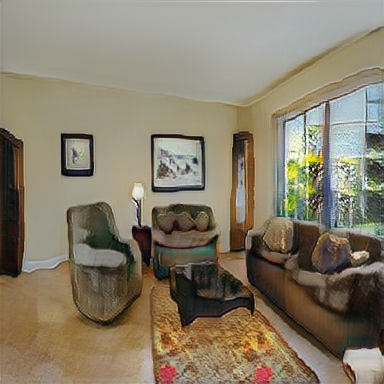}
\includegraphics[width=\fta, height=\fta]{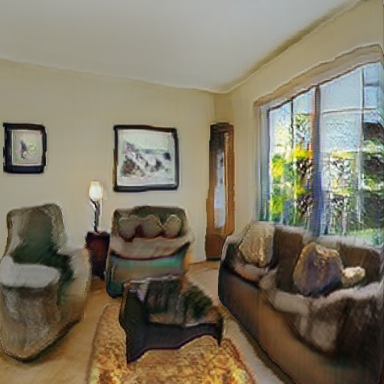} \hfill
\includegraphics[width=\fta, height=\fta]{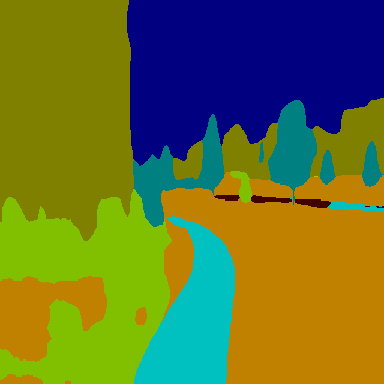}
\includegraphics[width=\fta, height=\fta]{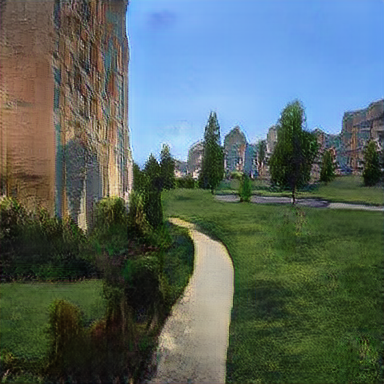}
\includegraphics[width=\fta, height=\fta]{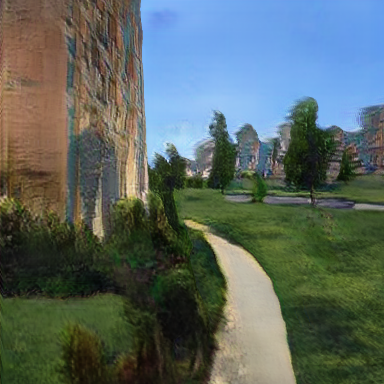}
\mpage{0.143}{\scriptsize Label}
\mpage{0.143}{\scriptsize Center view}
\mpage{0.143}{\scriptsize Novel view}
\hfill
\mpage{0.143}{\scriptsize Label}
\mpage{0.143}{\scriptsize Center view}
\mpage{0.143}{\scriptsize Novel view}
\caption{
    \textbf{Semantic view synthesis.} 
    We introduce a new visual synthesis problem, semantic view synthesis --- synthesizing a \emph{photorealistic} image that supports \emph{free-viewpoint rendering} given a single semantic label map.
    To achieve such visual effects, we build a two-step inference pipeline upon recent advances in semantic view synthesis and novel view synthesis.
    We show that our model learns to generate scene representations for rendering geometrically consistent and semantically meaningful novel views.
    We demonstrate the efficacy of our method using a wide variety of indoor (\emph{left}) and outdoor (\emph{right}) scenes.}
    \label{fig:teaser}
\end{figure}

%% file: 0_abstract.tex
\begin{abstract}
We tackle a new problem of semantic view synthesis --- generating free-viewpoint rendering of a synthesized scene using a semantic label map as input.
We build upon recent advances in semantic image synthesis and view synthesis for handling photographic image content generation and view extrapolation.
Direct application of existing image/view synthesis methods, however, results in severe ghosting/blurry artifacts.
To address the drawbacks, we propose a two-step approach.
First, we focus on synthesizing the color and depth of the visible surface of the 3D scene.
We then use the synthesized color and depth to impose explicit constraints on the multiple-plane image (MPI) representation prediction process.
Our method produces sharp contents at the original view and geometrically consistent renderings across novel viewpoints.
The experiments on numerous indoor and outdoor images show favorable results against several strong baselines and validate the effectiveness of our approach.
\end{abstract}

%% file: 1_introduction.tex
\section{Introduction}
\label{sec:intro}

Visual content creation using generative models has been gaining increasing attention.
Driving by the advances in generative models, recent work has demonstrated impressive performance on a wide range of tasks, including
image generation from various contexts (\eg noises~\cite{gan,stylegan}, images~\cite{pix2pix,cyclegan,drit,munit,albahar2019guided}, text~\cite{han2017stackgan,tseng2020retrievegan}, and audio~\cite{lee2019dancing2music}), 
view interpolation and extrapolation~\cite{deepstereo,hedman2018deep,stereomag,pushing,single_view_mpi},
and image editing~\cite{bau2019semantic,cheng2020rsegvae,tseng2020art}.
These algorithms greatly help unleash human imagination and support creative processes.
In this paper, we introduce a new form of visual content creation task by integrating 
(1) semantic image synthesis and (2) novel view synthesis. 

\emph{Semantic image synthesis}~\cite{crn,sims,pix2pixhd,spade} is a specific form of image-to-image translation task that aims to generate photorealistic images from semantic label maps.
Such an application is intuitive as users can easily draw and refine the semantic map on a digital canvas and then use the algorithm to synthesize \emph{2D images} with plausible appearances.
As these algorithms produce only 2D outputs, it is challenging for users to manipulate the viewpoints of the synthesized image in a geometrically consistent manner.  

\emph{View synthesis}, on the other hand, takes a sparse set of real images (captured at different viewpoints) as inputs and synthesizes novel views of the same scene~\cite{deepview,hedman2018deep,stereomag,pushing,single_view_mpi}.
This is achieved by explicitly or implicitly modeling the \emph{3D structure} of the scene.
However, these methods are applicable only to real images.

In this paper, we propose to tackle a new problem: \emph{semantic view synthesis} --- generating free-viewpoint rendering of a synthesized scene using a semantic label map as input (\figref{teaser}).
Compared to the existing semantic image synthesis task, the semantic view synthesis problem offers two unique advantages (\figref{application}).
First, it allows the users to easily manipulate the viewpoints of the synthesized image with minimal effort.
Second, it supports temporally and geometrically consistent rendering of 3D fly-through effects. 

To enable this new application, we develop a two-step method, drawing inspirations from the recent advances in semantic image synthesis and view synthesis algorithms.
First, given the input semantic label map, we leverage a state-of-the-art image synthesis model, SPADE~\cite{spade}, to generate a photorealistic color image and the corresponding disparity map.
The synthesized color/disparity images capture the appearance and structure of the \emph{visible surface} of the scene.
Second, to handle the dis-occluded contents (which become visible at novel views), we infer a multiplane images (MPI) representation~\cite{stereomag} using the synthesized color/disparity as constraints.
The resulting output of our method is an MPI representation that naturally supports view synthesis at any viewpoints.
We conduct extensive quantitative and visual comparisons on three datasets (ADE20K~\cite{ade}, ADE20k-outdoor~\cite{sims}, and NYUv2~\cite{nyu}) covering various indoor and outdoor scenes.
Our results demonstrate clear improvement over several strong baseline methods and alternative designs.

In summary, we make the following contributions:
\begin{itemize}
\item We introduce a new \textit{semantic view synthesis} task that aims to synthesize images of free-viewpoint from semantic masks. 
\item We propose a novel two-step training and inference pipeline: (1) color and disparity image synthesis for the visible surface and (2) MPI prediction with explicit constraints from the first step. (\secref{method})
\item We build several baseline approaches for this new problem and validate the efficacy of our proposed framework on a wide variety of indoor and outdoor scenes. (\secref{results})
\end{itemize}

\input{figure/fig_application}

%% file: figure/fig_application.tex
\begin{figure}[t!]
\centering
\includegraphics[width=\linewidth]{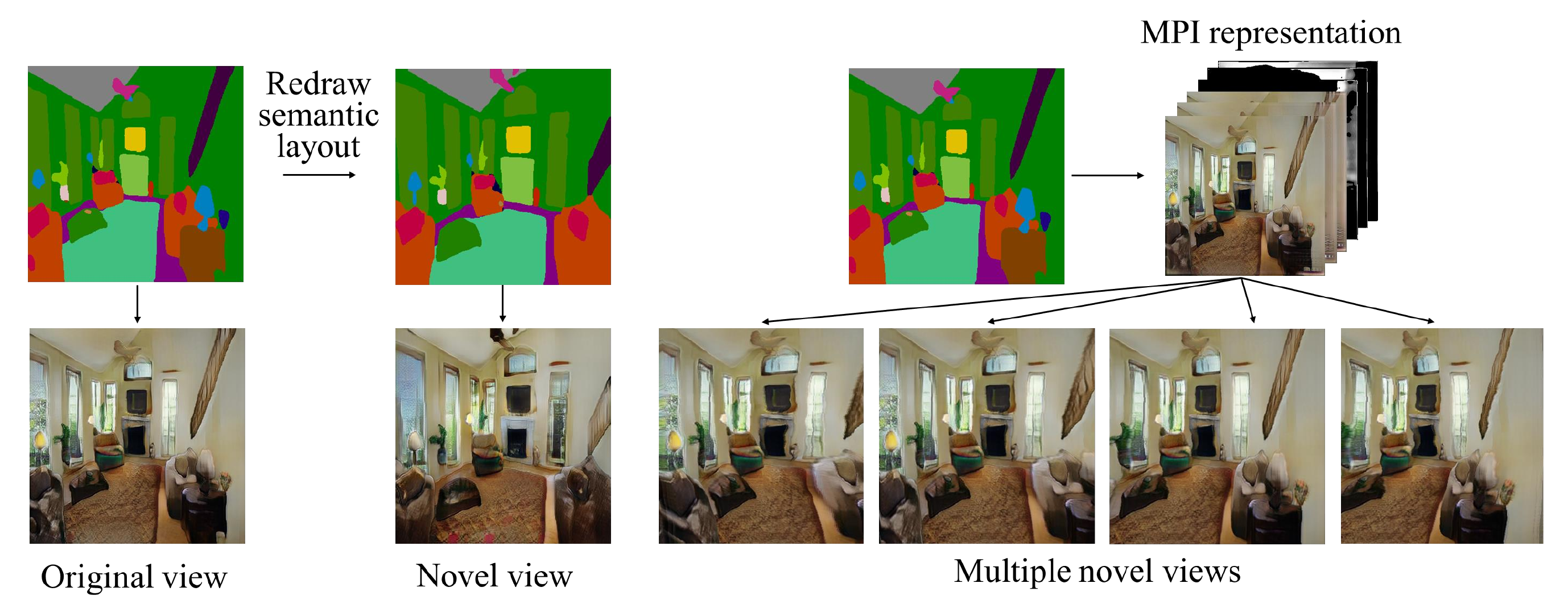}
\mpage{0.30}{\small Semantic image synthesis}
\mpage{0.60}{\small Semantic view synthesis (ours)}
\caption{\tb{Application.} The new problem of semantic view synthesis offers two advantages over the existing semantic image synthesis task. (a) \tb{Faster editing of viewpoints.}
(\emph{Left}) To refine the viewpoint of a synthesized image, the users would have to \emph{redraw} the semantic layout of the scene and apply the image synthesis algorithm on the new semantic layout again to produce the desired view.
(\emph{Right}) Taking a single semantic layout as input,  our method produces an MPI representation that naturally supports fast, free-form novel view rendering. 
(b) \tb{Consistent rendering over viewpoints.}
(\emph{Left}) As novel view images are \emph{independently} generated, the synthesized contents may not be consistent. 
(\emph{Right}) Our semantic view synthesis, in contrast, enables 3D fly-through effects with plausible motion parallax.
}
 
%
%
\label{fig:application}
\end{figure}

%% file: 2_prior.tex
\section{Related Work}
\label{sec:related}

\topic{Monocular depth prediction} aims to estimate the depth of a scene from a \emph{single-view} RGB image.
It is a challenging problem due to the difficulty of obtaining explicit 3D cue from the single-view RGB image without additional information~(\eg~stereo pair).
To conquer the problem, several supervised learning schemes~\cite{dorn,revisiting,fcrn,xu} utilize the ground-truth depth notation in the RGB-D dataset and train fully-convolutional networks (e.g., \cite{fcn}) to capture the image prior.
However, these approaches require large and diverse annotated data for the training.
Numerous self-supervised approaches~\cite{monodepth1,tinghui,dfnet,geonet,monodepth2} have been proposed to avoid the labor-intensive annotating process.
For instances, training with stereo videos \cite{monodepth1}, monocular videos \cite{monodepth2}, incorporating the information of camera poses or optical flow \cite{geonet,tinghui,dfnet,ZouLearning2020}. 
Nevertheless, these supervised and unsupervised methods often train their models using data from specific domains (e.g., driving scenes from the KITTI dataset) and therefore have difficulty in generalizing to diverse scenes in the wild.
On the other hand, a line of approaches uses multi-view internet photos \cite{megadepth}, MannequinChallenge~\cite{li2019learning} or 3D movies \cite{midas,wang2019web} as the source of data. 
In particular, training with mixed datasets from different sources achieves strong generality on unseen scenes. 
Our work leverage the pre-trained single-view depth estimation model from MiDaS \cite{midas} to obtain (pseudo) ground truth of depth/disparity maps for images in our training dataset.

\topic{Novel view synthesis} aims to generate novel views based on single or multiple images. 
Earlier learning-based approaches \cite{deepstereo,kalantari} take multiple posed images as input and produce the target views by blending the warped input images.
Such approaches, however, only \emph{interpolate} among the given viewpoints and do not handle dis-occlusions.
Recent advances explore generating novel view through a 3D scene representations, such as multi-plane images \cite{deepview,stereomag,pushing,locallf,single_view_mpi}, layered depth images \cite{peeking}, mesh representations \cite{3dphoto,Shih3DP20}, and point clouds \cite{synsin}. 
The multi-plane image representation \cite{deepview,locallf,pushing,single_view_mpi,stereomag} is a set of RGBA layers at discrete disparity levels. 
The novel views are rendered by homographic projection and alpha blending of the MPI layers.
The layered depth image approach \cite{peeking} represents 3D images as a foreground RGBD image and a background RGBD image. 
To generate the novel views, the RGB image is warped by the depth image, then composite by a predicted visibility mask. 
This approach requires supervision of the background image and only works for synthetic scenes.
3D photography \cite{3dphoto,Shih3DP20} focuses on generating 3D effects for real-world photos; they represent 3D images as a multi-layer 3D mesh.
These methods generate scene representation at the reference (original) viewpoint.
The novel view images can be rendered by projecting the scene representation to the desired viewpoint. 

Our work also produces an MPI representation as our output for supporting novel view synthesis.
Our problem setting, however, differs significantly from prior MPI-based methods.
Prior methods often require (at least) two images as inputs, which consist of the appearance of visible surfaces, cues of scene depth, and some content of the occluded background. 
In contrast, the input to our method is one semantic label map.
Our experimental results show that direct application of prior MPI-based methods leads to severe blurry ghosting artifacts when rendered at novel views.
Our two-step approach substantially reduces these artifacts via imposing explicit constraints on the MPI representation during training and testing time.



\topic{Image-to-image translation}
aims to learn the mapping between two image domains~\cite{albahar2019guided,munit,pix2pix,lee2019drit++,cyclegan,bicyclegan}. 
These techniques demonstrate a wide range of applications such as image inpainting, image super-resolution, domain adaptation \cite{chen2019crdoco,hoffman2018cycada}, and semantic image synthesis~\cite{pix2pixhd,spade}.
In particular, semantic image synthesis learns to generate photo-realistic images conditioned on semantic label maps.
Pix2pix~\cite{pix2pix} adopts a U-Net architecture to synthesize low-resolution images from a semantic map.
To operate in high-resolution settings, Pix2pixHD~\cite{pix2pixhd} introduces the multi-scale generator and discriminator network structure to enhance the quality of the generated images.
SPADE~\cite{spade} further improves Pix2pixHD with the spatially-adaptive normalization layers.
Different from the semantic image synthesis frameworks, we aim to synthesize 3D representation of a scene from a \emph{single-view} semantic segmentation layout.

\topic{Cross-modal distillation}
transfers the knowledge between different modalities. 
Existing works \cite{cross,modality} use learned representation from a large labeled dataset of the source modality as a supervised signal to train tasks of target modality with limited data.
For example, the method in \cite{cross} utilize ImageNet-pretrained model to train new representations for optical flow and depth images.
To address the problem of collecting a large indoor/outdoor dataset of semantic map to depth image pairs, our work also incorporates the idea of cross-modal distillation.
Specifically, We transfer the knowledge of monocular depth prediction model (predicting depth maps from images) and semantic segmentation (predicting semantic layouts from images) to our \emph{semantic depth synthesis} (predicting depth from semantic layouts).
To this end, we present a two-branch version of a SPADE network~\cite{spade} to predict both color and depth from a single semantic map.

%% file: 3_method.tex
\input{figure/fig_overview}

\section{Method}
\label{sec:method}

\subsection{Overview}
\label{sec:overview}
Our goal is to learn to synthesize novel-view color images from a given a semantic label map.
As shown in~\figref{overview}, our scene representation generation process consists of (1) image and disparity generation module and (2) MPI prediction module.
With the generated MPI, we can project and blend the MPI to produce the desired target views.
%
In this section, we first describe the data preparation in~\subsecref{data}.
We then detail the training procedure of scene representation generation including image and disparity generation and the MPI prediction in~\subsecref{scene}.
Finally, we introduce the novel view synthesis procedure at test time in~\subsecref{novel}.

\subsection{Data preparation}
\label{subsec:data}
We build a dataset from the RealEstate10K dataset \cite{stereomag}, which consists of 80,000 indoor/outdoor YouTube video clips with camera poses for each frame.
To extract training pairs of the semantic layout and the corresponding disparity map, we adopt the idea of cross-modal distillation (\figref{method}a).
Specifically, we apply PSPNet \cite{pspnet} (pretrained on the ADE20K~\cite{ade}) to obtain segmentation map annotation.
Similar, we apply the pre-trained MiDaS \cite{midas} monocular depth estimation network to estimate the corresponding disparity map. 
Since MiDaS predicts the \emph{relative} disparity with unknown scale/shift, we use the absolute depth prediction from DPSNet \cite{dpsnet} to estimate the scale and shift for each training image.
The relative disparity images are then transformed into absolute disparity images that serve as the (pseudo) ground-truth images for training.
We collect training pairs from each frame in the RealEstate10K dataset.
While existing Habitat \cite{habitat} framework also provides semantic layouts, disparity maps and multi-view images with camera poses, we did not use it as the dataset contains indoor scenes only. 

\subsection{Scene Representation Generation}
\label{subsec:scene}
We adopt a two-step prediction strategy due to the difficulty of predicting MPI representation in one step.
First, our image and disparity generator takes the semantic layout $l$ as input and learns to synthesize the corresponding color image $\hat x^s_{FG}$ and disparity image $\hat{d}^s$ of the visible surface.
Second, the MPI generator uses the synthesized color image and the disparity as input and predicts an MPI representation $\hat m^s$ of the scene.

\input{figure/fig_depth}

\topic{Image and disparity generation.}
\label{subsubsec:twostream}
Image and disparity generator aims to synthesize the color $\hat x^s_{FG}$ and disparity image $\hat d^s$ of  \emph{visible surface} of the scene (\figref{method}b).
To this end, we modify the SPADE~\cite{spade} model into two-stream generators (with the color generator $G_{x}$ and the disparity generator $G_{d}$).
The two-stream generators $G_{x}$ and $G_{d}$ share the first three SPADE-style ResNet blocks.
%
Using the training pairs of semantic layout $l$ and disparity image $d$, we use the losses in SPADE~\cite{spade} for training the color stream and an $\ell1$ reconstruction loss for training the disparity stream.
\figref{depth} shows sample results of disparity prediction from a semantic label map.


\input{figure/fig_method}

\topic{MPI prediction.}
\label{subsubsec:mpi}
For simple scenes (e.g., there is no apparent occluded region in the input image), using a single image with the associated disparity map will suffice for modeling the 3D scene.
However, synthesizing novel-view images with only color and disparity map inevitably induce visible artifacts, particularly in the dis-occluded regions, thereby failing to render general scenes where multiple depth layers exist.
We therefore use an MPI representation~\cite{stereomag} for handling the depth-complex scenarios.
An MPI~\cite{stereomag} $m=\{(x_k,\alpha_k)\}^{K}_{k=1}$ is a collection of RGBA images, where $K$ is the number of depth planes.
Each layer $k$ is an image plane placed at a fixed depth with respect to a virtual reference camera.
The color images ${x_k}$ at each depth plane indicate the visible view, while the alpha image $\alpha_k$ represents the visibility, which has a range between $0$ and $1$.

\input{figure/fig_blur}

However, we find that predicting the MPI using only a single color image results in poor visual quality. 
The primary reason is that without depth cues (e.g., stereo pair in \cite{stereomag}), it is challenging to predict accurate alpha (transparency) maps for compositing multi-plane images. 
To tackle this issue, we directly compute and constrain the alpha images from the synthesized disparity map $\hat d^s$.
Since the synthesized disparity map $\hat d^s$ provides a strong prior for the scene visibility at different depth layers, we transform it into the alpha images $\{\hat \alpha^s_{k}\}$ in our MPI representation (\figref{blur}).
Specifically, we first transform the disparity image into a \emph{one-hot representation} with K disparity channels, according to the inverse depth.
Then, we apply a half Gaussian blur along the disparity channel, which produces blurring effect only \emph{behind} the predicted disparity and has a peak value at the predicted disparity.
The blurred one-hot disparity images are then used as the alpha images in our MPI representation.

The alpha images generated by this simple process has three desired properties. 
First, the pixels at the predicted disparity level are fully visible, resulting in sharp contents at the center view. 
Second, the blurred alpha images allow the MPI generator to predict the BG colors and blending weights for handling dis-occluded regions at novel views.
Third, as the alpha images are generated in a deterministic manner, the MPI generator can focus only on predicting the color images at multiple planes.

%

To predict the color images, $\{\hat x^s_{k}\}$ in the MPI representation, we use a SPADE-based \cite{spade} MPI generator $G_{m}$ that takes the color image of the visible surface $\hat x^s_{FG}$ as main input, and uses the disparity image  $\hat d^s$ for modulating the activations in normalization layers. The MPI generator synthesizes a background color image $\hat x^s_{BG}$ and a set of blending weights $\{\hat w_{k}\}$. 
The color images $\{\hat x^s_{k}\}$ are calculated as the weighted sum of the foreground $\hat x^s_{FG}$ and the background $\hat x^s_{BG}$: 
\begin{equation}
\hat x^s_{k} = \hat w_{k} \odot \hat x^s_{FG} + (1- \hat w_{k}) \odot \hat x^s_{BG} 
\label{eq:blend}
\end{equation}
We refer the reader to Zhou~\etal~\cite{stereomag} for more details on synthesizing novel view images using an MPI representation. 


\topic{Training MPI generator.} 
\figref{method}c illustrates the training process of MPI prediction.
We use the data sampling strategy in \cite{stereomag} to sample the training image pair $(x^s, x^n) = (x^s_{FG}, x^n)$ (note that $x^s$ is equivalent to $x^s_{FG}$) with corresponding camera poses $(p^s, p^n)$ , as well as the disparity image $d^s$, where the notation $s$ and $n$ indicate the \emph{source} and \emph{novel} view, respectively.
Our MPI generator predicts the color images $\{\hat x^s_{k}\}$ from the source color image $x^s_{FG}$. 
We transform the disparity image $d^s$ into alpha images $\{\alpha^s_{k}\}$. 

With the predicted MPI representation $\hat m^s = (\{\hat x^s_{k}\}, \{\alpha^s_{k}\})$, we can use the warped multi-plane images according to the relative pose $p^{n-s}$ between the source pose $p^s$ and novel pose $p^n$.
Given the warped MPIs, we then use the over-composited approach~\cite{over} to composite the novel view $\hat x^n$.
We train the MPI generator using an $\ell1$ loss and a GAN loss of weight 0.01 between the generated and the ground-truth color image at the novel view $x^n$.
%

\subsection{Novel view synthesis}
\label{subsec:novel}

Similar to the training process, at test time, we follow the two-step approach for generating an MPI.
First, we generate color $\hat{x}^s_{FG}$ and disparity image $\hat{d}^s$ from input semantic layout $l$.
We then use both color $\hat{x}^s_{FG}$ and disparity image $\hat{d}^s$ to predict the MPI representation $\hat m^s = (\{\hat x^s_{k}\}, \{\hat \alpha^s_{k}\})$. 
Given a relative camera pose, we can warp and over-composite the predicted MPI and obtain the novel view image $\hat{x}^n$.


%% file: figure/fig_overview.tex
\begin{figure*}[t]
\includegraphics[width=\linewidth]{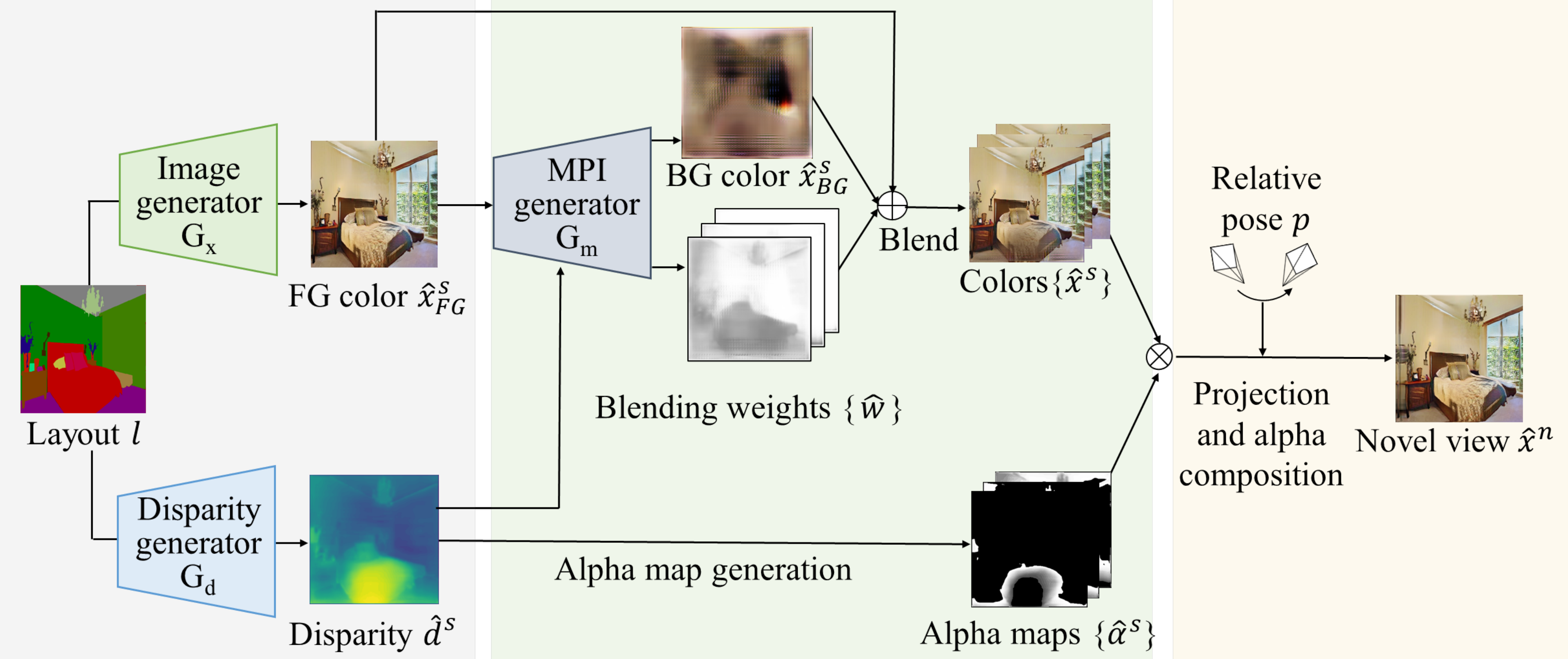} \\
\mpage{0.25}{\footnotesize (a) Image/disparity generation (Sec. \ref{subsubsec:twostream})}
\mpage{0.45}{\footnotesize (b) MPI prediction (Sec. \ref{subsubsec:mpi})}
\mpage{0.25}{\footnotesize (c) View synthesis (Sec. \ref{subsec:novel})}
\caption{
\textbf{Method overview.} 
Our method first produces an MPI-based scene representation via a two-step approach (a)(b). 
(a) Our first step focuses on synthesizing the color and disparity image from the given semantic label map as the \emph{visible surface}.
Here, we present a Y-shaped network with partially shared color/depth decoder architecture to ensure consistency between the synthesized color and depth maps. 
(b) We then infer the MPI representation that captures the color and structure of both the visible surface and the dis-occluded surfaces.
With only one single RGB image as input, it is challenging to learn MPI with high-quality view renderings.
This is because the network needs to predict both the appearances at multiple depth levels as well as the alpha (transparency) maps.
To address this issue, we directly generate the alpha maps using the synthesized depth map from step (a), and we use the synthesized depth map for modulating the activations in normalization layers \cite{spade} in our MPI generator.
Such an approach imposes effective constraints and results in improved MPI prediction. 
(c) Given target camera poses, we can then project and blend the generated MPI representation for rendering images at novel views.
}
\label{fig:overview}
\end{figure*}

%% file: figure/fig_depth.tex
\begin{figure}[t!]
    \centering
    \includegraphics[width=0.115\linewidth, height=0.115\linewidth]{figure/fig_depth/rgb/0522.png}
    \includegraphics[width=0.115\linewidth, height=0.115\linewidth]{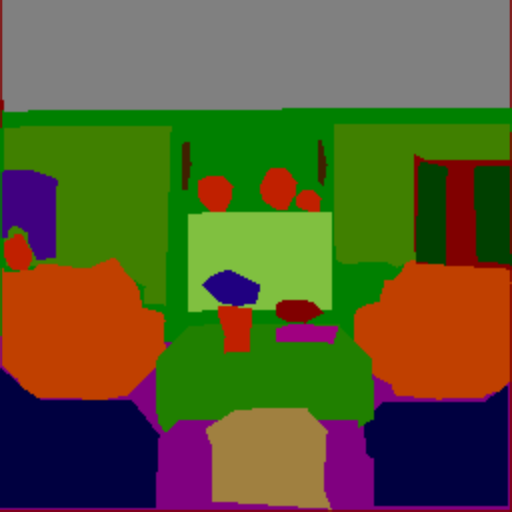}
    \includegraphics[width=0.115\linewidth, height=0.115\linewidth]{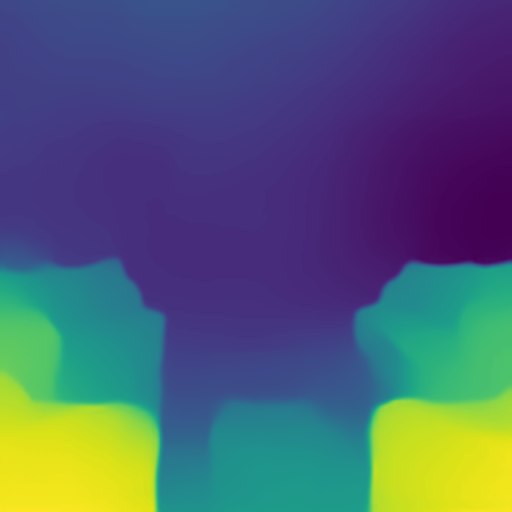}
    \includegraphics[width=0.115\linewidth, height=0.115\linewidth]{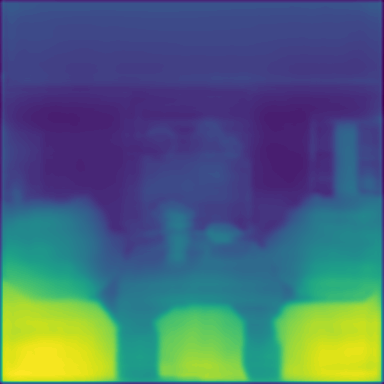}
    \space
    \includegraphics[width=0.115\linewidth, height=0.115\linewidth]{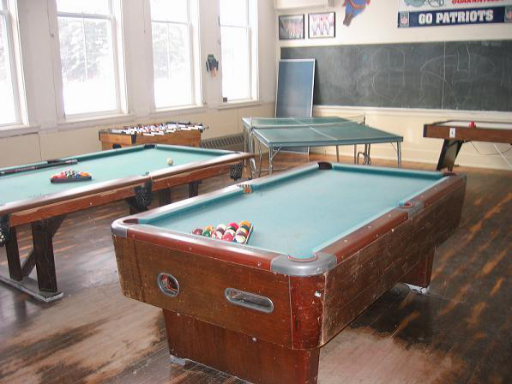}
    \includegraphics[width=0.115\linewidth, height=0.115\linewidth]{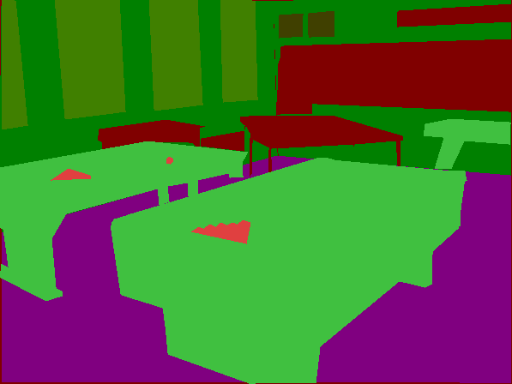}
    \includegraphics[width=0.115\linewidth, height=0.115\linewidth]{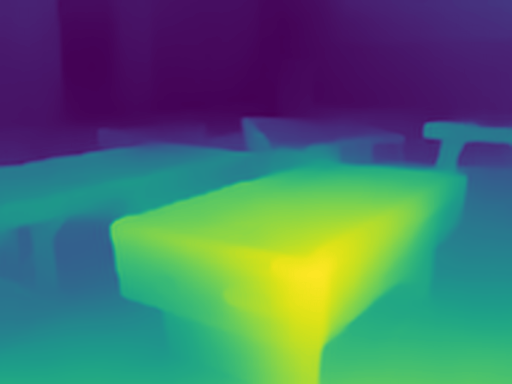}
    \includegraphics[width=0.115\linewidth, height=0.115\linewidth]{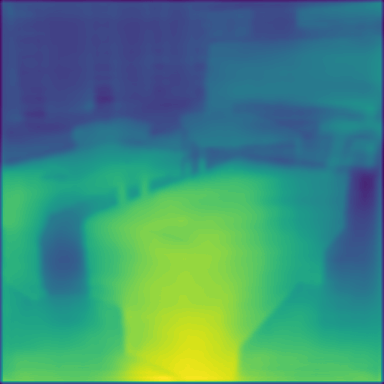} \\


    \includegraphics[width=0.115\linewidth, height=0.115\linewidth]{figure/fig_depth/rgb/0002.png}
    \includegraphics[width=0.115\linewidth, height=0.115\linewidth]{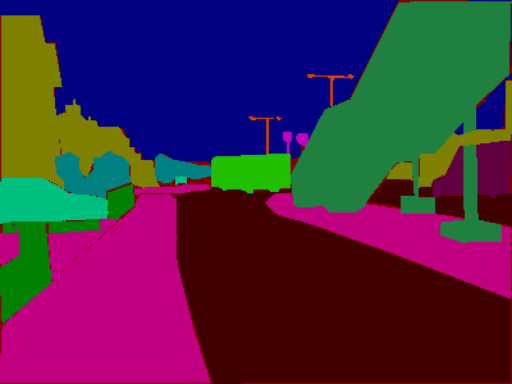}
    \includegraphics[width=0.115\linewidth, height=0.115\linewidth]{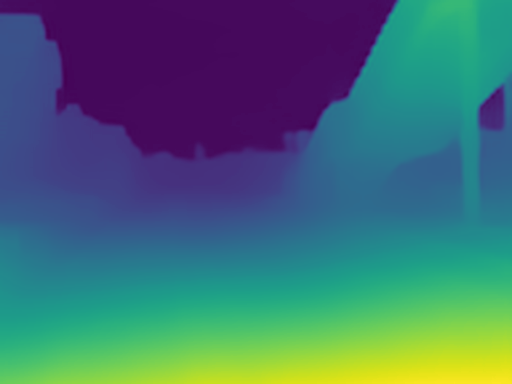}
    \includegraphics[width=0.115\linewidth, height=0.115\linewidth]{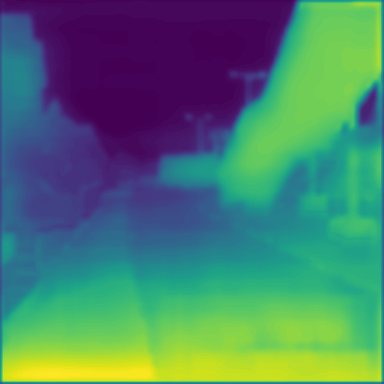}
    \space
    \includegraphics[width=0.115\linewidth, height=0.115\linewidth]{figure/fig_depth/rgb/0542.png}
    \includegraphics[width=0.115\linewidth, height=0.115\linewidth]{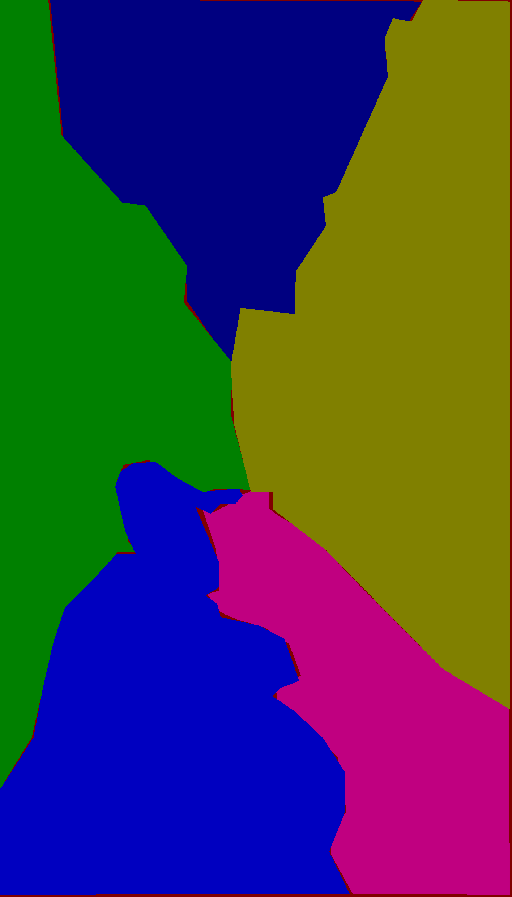}
    \includegraphics[width=0.115\linewidth, height=0.115\linewidth]{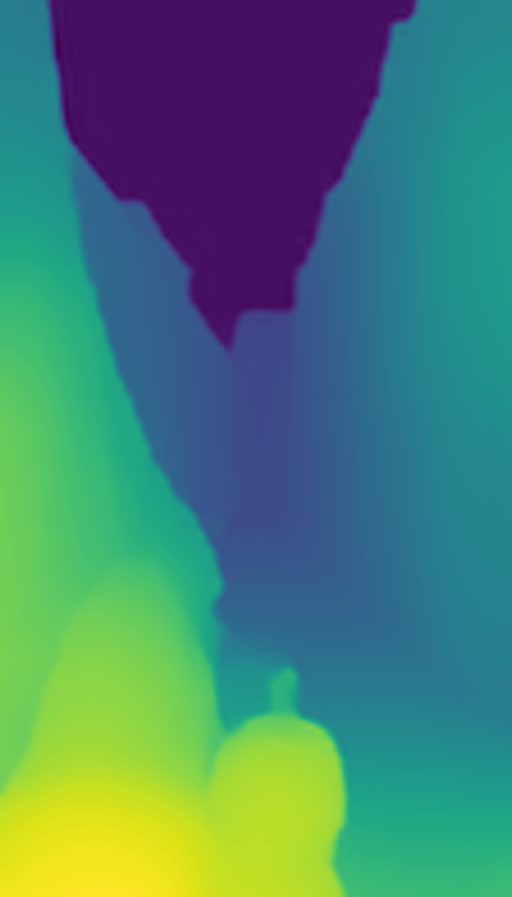}
    \includegraphics[width=0.115\linewidth, height=0.115\linewidth]{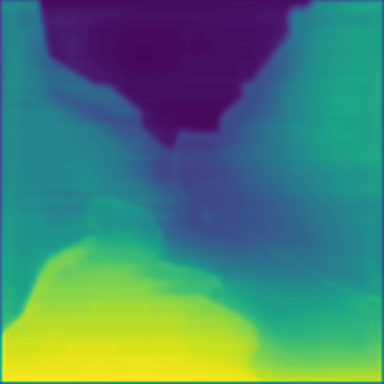}\\

    \mpage{0.10}{\scriptsize Image}
    \mpage{0.10}{\scriptsize Label}
    \mpage{0.10}{\tiny MiDaS \cite{midas} \\ (from image)}
    \mpage{0.10}{\tiny Our results \\ (from label)}
    \space
    \mpage{0.10}{\scriptsize Image}
    \mpage{0.10}{\scriptsize Label}
    \mpage{0.10}{\tiny MiDaS \cite{midas} \\ (from image)}
    \mpage{0.10}{\tiny Our results \\ (from label)}

    \captionof{figure}{\textbf{Sample results of depth synthesis.} Comparing the prediction from MiDaS~\cite{midas} (computed from color images), our model produces plausible depth images based on semantic label maps.
    }
    \label{fig:depth}
\end{figure}

%% file: figure/fig_method.tex
\begin{figure*}[t!]
\centering
\includegraphics[width=\linewidth]{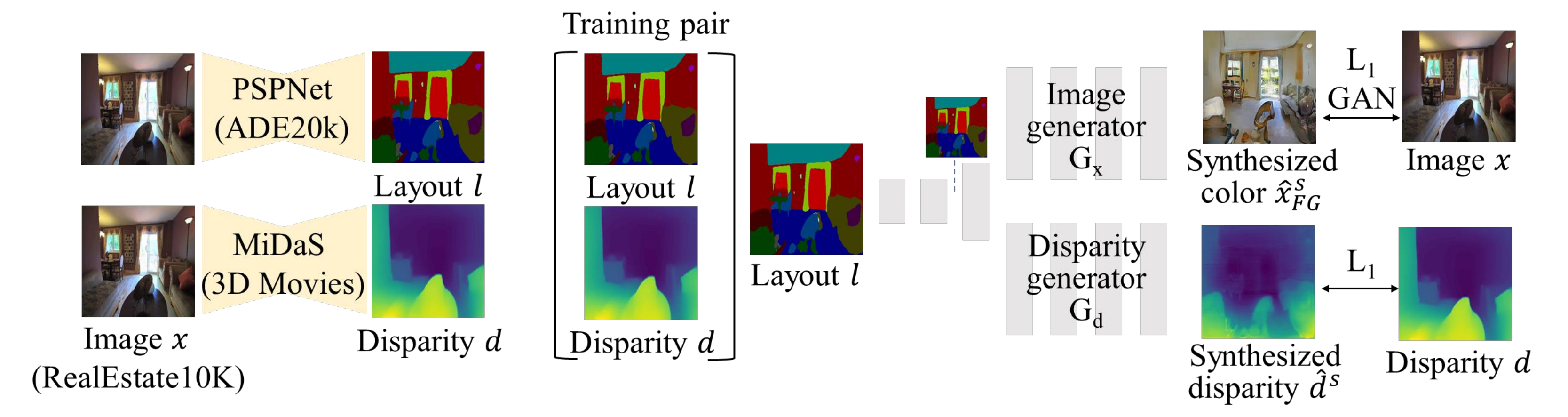}
\mpage{0.4}{\footnotesize  (a) Cross-modal distillation} \hfill
\mpage{0.55}{\footnotesize  (b) Semantic image / disparity synthesis}

\includegraphics[width=\linewidth]{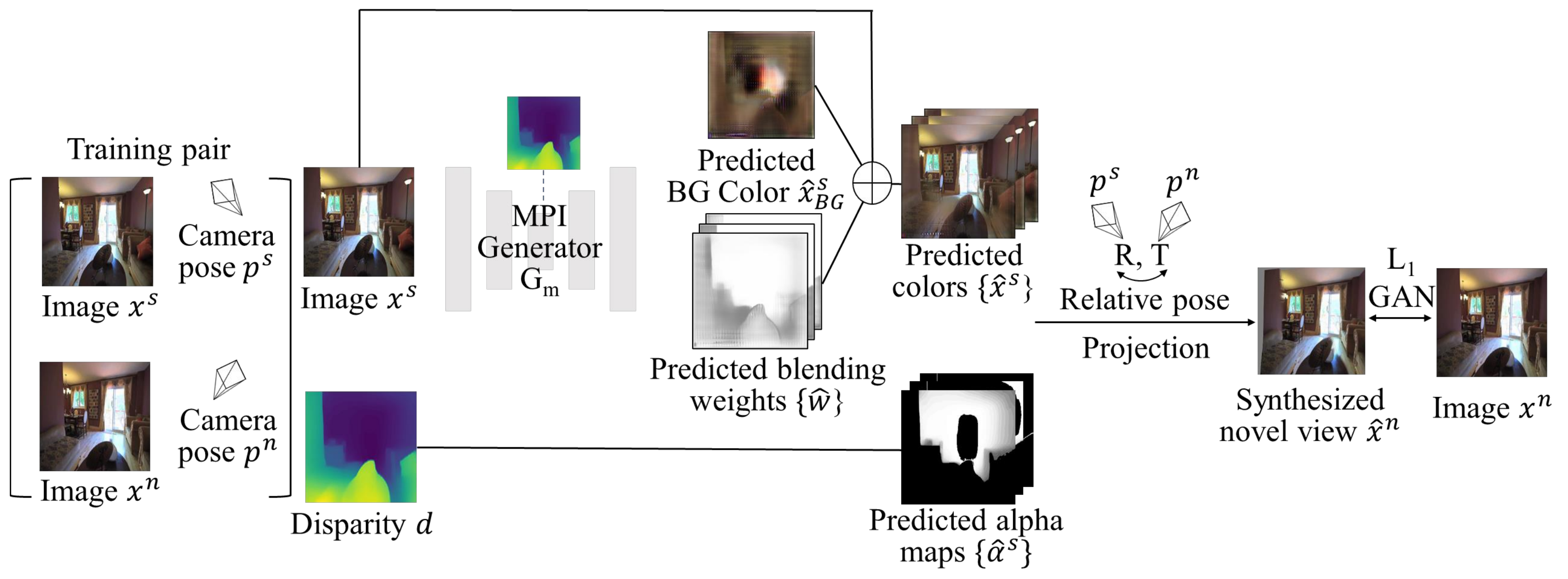}
\mpage{1}{\footnotesize  (c) MPI prediction}
\caption{\textbf{Training pipeline.} 
Our model training process consists of the following steps.
(a) \textbf{Cross-modal distillation}: 
We generate \emph{pseudo} training pairs for training the semantic image/depth generation by applying the pre-trained depth estimation model~\cite{midas} and semantic segmentation PSPNet~\cite{pspnet} on training images.
(b) \textbf{Semantic depth and image synthesis}: Using the generated training pairs from the cross-modal distillation step, we use a two-stream (color and disparity) SPADE network to generate the visible surface.
%
%
We train the color stream using the losses in \cite{spade} and the disparity stream with an $\ell1$ loss (based on the normalized disparity values). 
%
(c) \textbf{MPI prediction}: 
We use training pairs of source/target images with relative pose annotations (provided by \cite{stereomag}).
We train the MPI generator to produce colors at multiple depth levels and use $\ell1$ and GAN loss to enforce the consistency between the projected image and the target image. 
Note that the MPI generator does \emph{not} need to predict the alpha (transparency) maps.
%
%
%
%
} 
\label{fig:method}
\end{figure*}

%% file: figure/fig_blur.tex
\begin{figure}[t!]
\centering
\includegraphics[width=\linewidth]{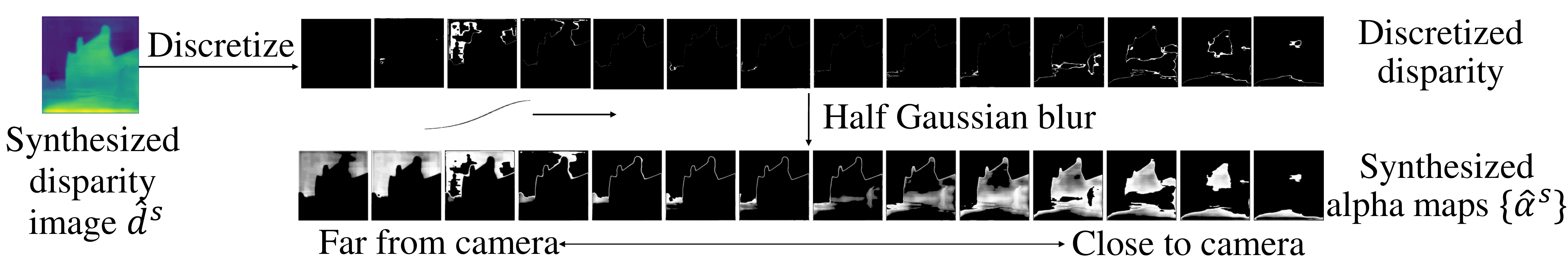} \\
\mpage{0.25}{\small (a) Discretization}\hfill
\mpage{0.5}{\small (b) Gaussian blur} \hfill
\caption{\textbf{Alpha images.} (a) \textbf{Discretization}: we first transform the disparity image to a one-hot encoding image. (b) \textbf{Gaussian blur}: we then apply a half Gaussian blur along the disparity channel, and use the result as our alpha images. The alpha images shown here are 14 out of total 128 planes.
}
\label{fig:blur}
\end{figure}

%% file: 4_result.tex
\section{Experimental Results}
\label{sec:results}

\subsection{Experimental setup}

\topic{Datasets.}
We validate our method on three datasets. 
\begin{compactitem}
\item{\bf ADE20K}~\cite{ade} is a dataset of diverse indoor and outdoor scenes. It consists of 2,000 testing images with 150 semantic classes.

\item{\bf ADE20K-outdoor}~\cite{sims} is a subset of outdoor scenes in ADE20K dataset. It consists of 1,035 testing images with 150 semantic classes. 

\item{\bf NYU}~\cite{nyu} is an indoor dataset. It consists of 249 testing images with 13 semantic classes. 
\end{compactitem}

\topic{Implementation details.} 
We implement our system in PyTorch and use the Adam optimizer with $\beta_1 = 0$, $\beta_2 = 0.9$ for all network training.
All the experiments are conducted on an NVIDIA GTX 1080.
The color module, the disparity module and the MPI module are trained for 600k/300k/300k iterations respectively. 
We use a batch size of one with a learning rate of 0.0002.
We use $K=128$ image planes for our MPI representations. 
We set the disparity of each alpha map equally distributed from 0.01m to 1m, according to the inverse depth. 
The Gaussian blur we use for the alpha images has a peak 1, window 31, and the $\sigma$ value of 10. 
We set the size of the target synthesized images as $384 \times 384$ for all the models. 
Our source code and the pre-trained models are available on the project website.

\topic{Baselines.} 
We compare our methods with four baseline methods. 
\begin{compactitem}
\item{(a) \textbf{Direct (U-Net)}} synthesizes the multi-plane images directly from the semantic layout using a fully-convolutional encoder-decoder architecture~\cite{stereomag}.

\item{(b) \textbf{Direct (SPADE)}} also synthesizes the multi-plane images directly from the semantic layout, but uses a generator with spatially-adaptive normalization~\cite{spade}.

\item{(c) \textbf{Cascade (MPI)}} first synthesizes a color image from the semantic layout using SPADE \cite{spade}, then apply an MPI predictor using the synthesized image as input.
Here, we modify the original MPI generation model in \cite{stereomag} so that it takes a single image as input.

\item{(d) \textbf{Cascade (KB)}} first synthesizes a color image from the semantic layout using SPADE \cite{spade}, then apply a recent single-image view synthesis method (3D Ken Burns \cite{ken}).
\end{compactitem}
Training and testing details of the baseline models can be found in the supplementary material.

\subsection{Quantitative evaluation}

\input{figure/fig_fid}

We use the Fr\'echet Inception Distance (FID) \cite{FID} to measure the distance between the distribution of generated images and real images.
We use ADE20K images as real images.
For measuring the realism of novel view synthesis, we evaluate the FID scores of generating novel views at $7 \times 7$-grid viewpoints on x-y planes with camera movement from $-0.3$ meter to $0.3$ meter across both axes. 
The center view with camera movement $(0,0)$ shows the performance of semantic image synthesis.
%
As shown in \figref{fid}, all the baselines, and our model produce the lowest FID score at the center view, and the FID score gradually increases when the camera movement becomes larger. 
The trend is similar across different datasets. 
We discuss the results based on the ADE20K dataset below.

\topic{Results at the center view.}
Comparing methods directly synthesizing MPIs from layouts, {Direct (SPADE)} performs better than {Direct (U-Net)} (102 vs. 128) due to the use of the SPADE architecture. 
Comparing methods that both employ the SPADE generator, {Cascade (MPI)} performs better than {Direct (SPADE)} (50 vs. 102), suggesting the difficulty of directly predicting MPI from semantic layout. 
Our method achieves the same FID score 50 when compared with {Cascade (MPI)} at the center view as the input (synthesized color image) is the same.

\topic{Results at the novel views.}
When evaluating the results at a novel view (e.g., $(0.3, 0.3)$ meters away from the center), we observe that while the {Cascade (MPI)} method performs well at the center view, it produces significantly inferior to the methods that directly predict MPI.
In contrast, our method produces lowest FID scores among the competing baselines. 


\input{figure/fig_result}

\subsection{Visual comparisons}
\figref{result} compares the generated novel view images of four baselines and our model. 
Two-step methods, {Cascade (MPI)}, {Cascade (3D Ken Burns)} and {Ours}, produce images with sharper contents. 
{Direct (U-Net)} and {Direct (SPADE)} tend to produce blurry and less plausible contents. 
In particular, the results of {Cascade (MPI)} suffer from blurry due to the difficulty of generating alpha images when no depth cues (\eg multiple images, plane sweep volume) are available. 
The {Cascade (KB)} inpaints the dis-occluded region at only \emph{one} novel viewpoint.
Such a method supports 3D Ken Burns effect with a simple camera trajectory such as zooming in, but not free-viewpoint rendering.


%

\input{figure/tab_ablation}

\subsection{Ablation study}

\input{figure/fig_ablation}

\topic{Number of depth layers.} 
\tabref{ablation}a shows the results of having a different number of depth layers in our MPI. 
At $(0.2, 0.2)$, the model with $K=32$ achieves better FID.
At $(0, 0)$ and $(0.1, 0.1)$, the model with $K=128$ achieves better FID. 
We conclude that more MPI planes lead to slightly blurrier results for large camera movement.
%
\figref{ablation} illustrates that the novel view synthesized with 32 depth layers show more artifacts than 64 or 128 depth layers.

\input{figure/fig_dis}

\topic{Background inpainting.} 
We explore alternative methods for handling the dis-occluded regions when rendering at novel views.
We use the standard backward warping to project the synthesized color image using disparity image to render the novel views.
We then inpaint the missing pixels using either simple diffusion (implemented in OpenCV) or a learning-based image inpainting model (GatedConv~\cite{gate}).

\tabref{ablation}b shows that our method achieves lower FID scores at three viewpoints.
Note that as all the novel view images are processed independently, \emph{Diffusion} and \emph{GatedConv} approaches do not retain the consistency across different viewpoints.
We refer the readers to the supplementary materials for video results. 
\figref{dis} shows that while our method produces slightly blurry foreground (due to the over-composition of multi-plane images), our MPI representation hallucinates plausible dis-occluded regions.

\input{figure/fig_user}

\subsection{User study}
We conducted a perceptual user study to quantify the user preference over the proposed method and the six baseline approaches. 
For each test during the study, we present two novel view videos of the same scene generated by two different methods with circular camera motion (in randomized order). We then ask the participant to select his/her preferred result. 
There are 120 videos (60 pairwise comparisons) generated from the layouts in ADE20K, ADE20K-outdoor, and NYU datasets used. 
We conduct the study with 47 participants (2820 binary votes). 
The results shown in \figref{user} validate that the proposed method synthesizes more realistic novel view videos compared to the baseline approaches.

%% file: figure/fig_fid.tex
\begin{figure}[t!]
\begin{center}
    \centering

    \mpage{0.02}{\raisebox{0pt}{\rotatebox{90}{\footnotesize ADE20K}}} 
    \mpage{0.215}{\includegraphics[width=\linewidth, trim={0.4cm 0.5cm 0.3cm 0.5cm},clip ]{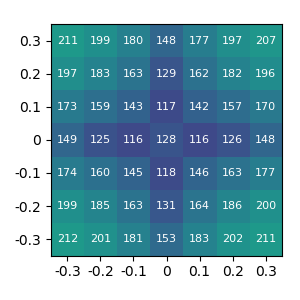}}
    \mpage{0.215}{\includegraphics[width=\linewidth, trim={0.4cm 0.5cm 0.3cm 0.5cm},clip ]{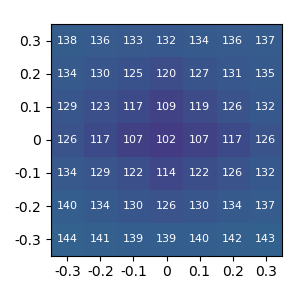}}
    \mpage{0.215}{\includegraphics[width=\linewidth, trim={0.4cm 0.5cm 0.3cm 0.5cm},clip ]{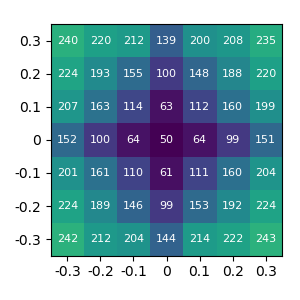}}
    \mpage{0.215}{\includegraphics[width=\linewidth, trim={0.4cm 0.5cm 0.3cm 0.5cm},clip ]{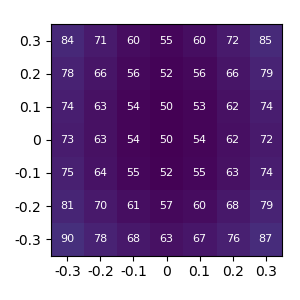}} \\
    \mpage{0.02}{\raisebox{0pt}{\rotatebox{90}{\footnotesize ADE20K-outdoor}}} 
    \mpage{0.215}{\includegraphics[width=\linewidth, trim={0.4cm 0.5cm 0.3cm 0.5cm},clip ]{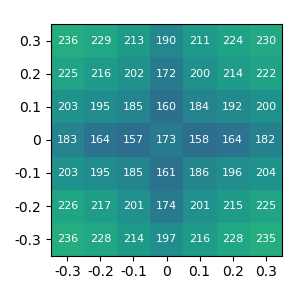}}
    \mpage{0.215}{\includegraphics[width=\linewidth, trim={0.4cm 0.5cm 0.3cm 0.5cm},clip ]{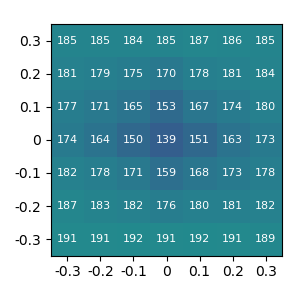}}
    \mpage{0.215}{\includegraphics[width=\linewidth, trim={0.4cm 0.5cm 0.3cm 0.5cm},clip ]{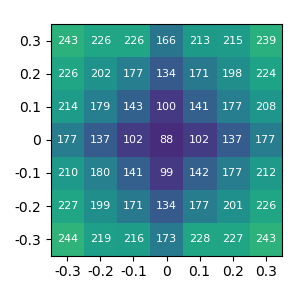}}
    \mpage{0.215}{\includegraphics[width=\linewidth, trim={0.4cm 0.5cm 0.3cm 0.5cm},clip ]{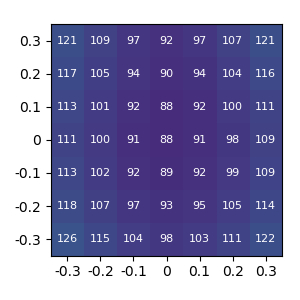}} \\
    \mpage{0.02}{\raisebox{0pt}{\rotatebox{90}{\footnotesize NYU}}} 
    \mpage{0.215}{\includegraphics[width=\linewidth, trim={0.4cm 0.5cm 0.3cm 0.5cm},clip ]{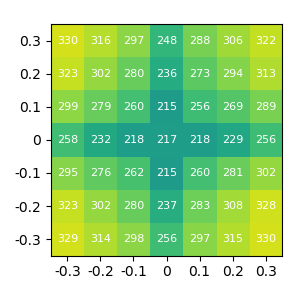}}
    \mpage{0.215}{\includegraphics[width=\linewidth, trim={0.4cm 0.5cm 0.3cm 0.5cm},clip ]{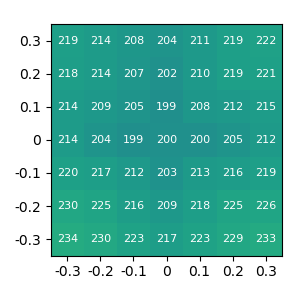}}
    \mpage{0.215}{\includegraphics[width=\linewidth, trim={0.4cm 0.5cm 0.3cm 0.5cm},clip ]{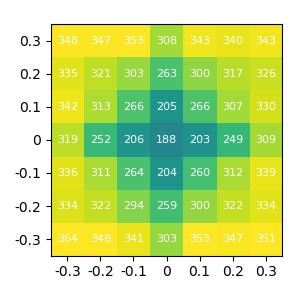}}
    \mpage{0.215}{\includegraphics[width=\linewidth, trim={0.4cm 0.5cm 0.3cm 0.5cm},clip ]{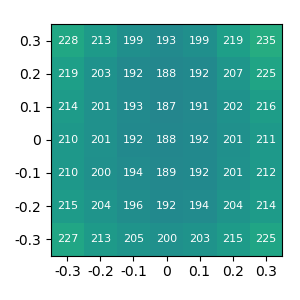}} \\

    \mpage{0.02}{\raisebox{0pt}{\rotatebox{90}{}}}
    \mpage{0.215}{\small Direct (U-Net)}
    \mpage{0.215}{\small Direct (SPADE)}
    \mpage{0.215}{\small Cascade (MPI)}
    \mpage{0.215}{\small Ours}

    \captionof{figure}{\textbf{Quantitative evaluation.}
We compare the results of three alternative approaches for semantic view synthesis and our model on the ADE20K, ADE20K-outdoor, and NYU datasets. 
Each table shows the FID score of generated novel view images at $7\times 7$ grids of target viewpoints. 
A lower FID score is better.
Using {Cascade (MPI)} and {Direct (U-Net)} MPI architectures is unable to produce sharp, photorealistic contents (therefore high FID scores). 
The {Direct (SPADE)} method can synthesize detailed contents at the center view due to the use of SPADE~\cite{spade}. 
However, its performance degrades rapidly when the camera viewpoints move away from the center view. 
Our two-step generation preserves the detailed content in the front layer while maintaining photorealism under novel views. 
We were not able to include Cascade (KB) due to different camera movements.
    }
    \label{fig:fid}
\end{center}
\end{figure}

%% file: figure/fig_result.tex
\newlength\ftb
\setlength\ftb{0.155\linewidth}

\begin{figure}[t!]
\begin{center}
    \centering

    \includegraphics[width=\ftb, height=\ftb,clip,trim={25px 0 0 25px}]{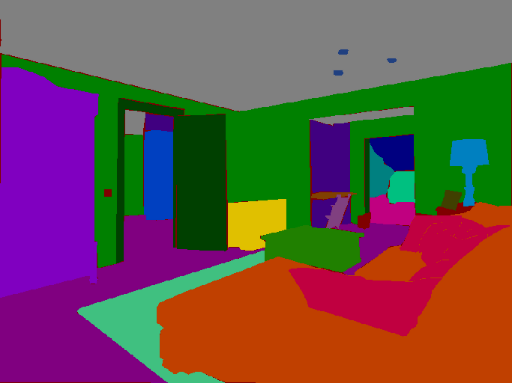} \hfill
    \includegraphics[width=\ftb, height=\ftb,clip,trim={25px 0 0 25px}]{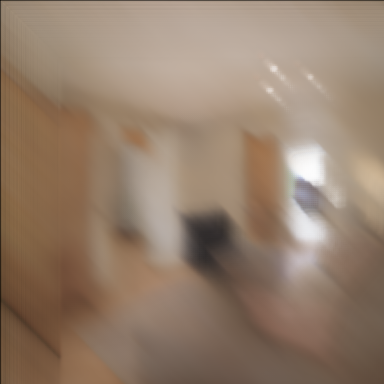} \hfill
    \includegraphics[width=\ftb, height=\ftb,clip,trim={25px 0 0 25px}]{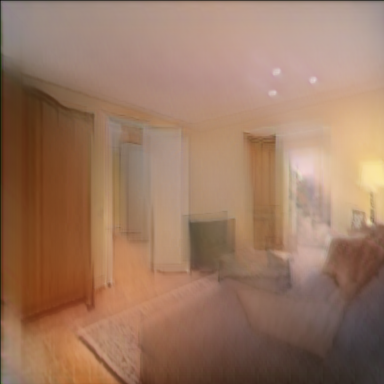} \hfill
    \includegraphics[width=\ftb, height=\ftb,clip,trim={25px 0 0 25px}]{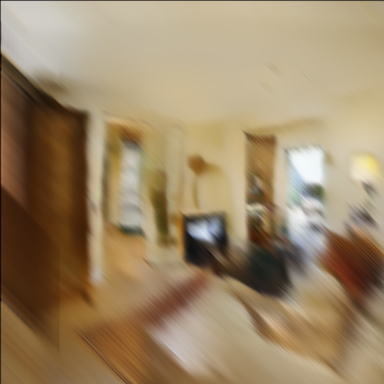} \hfill
    \includegraphics[width=\ftb, height=\ftb,clip,trim={25px 0 0 25px}]{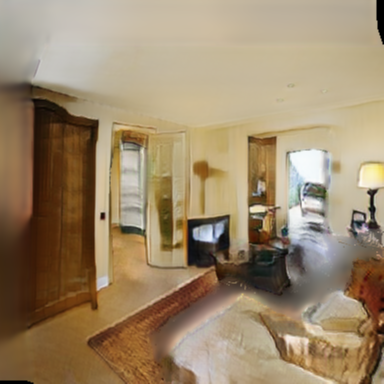} \hfill
    \includegraphics[width=\ftb, height=\ftb,clip,trim={25px 0 0 25px}]{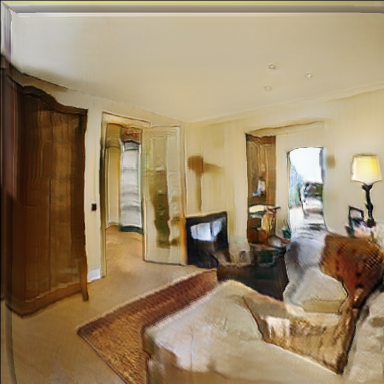}   \\
																
											
    \includegraphics[width=\ftb, height=\ftb,clip,trim={25px 0 0 25px}]{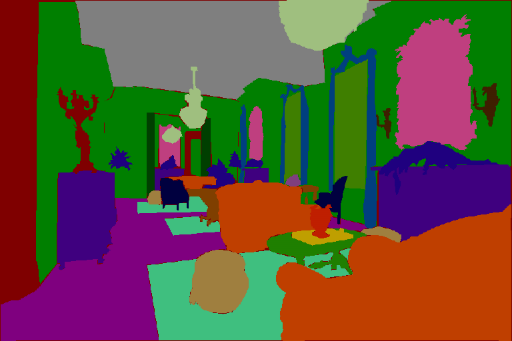} \hfill
    \includegraphics[width=\ftb, height=\ftb,clip,trim={25px 0 0 25px}]{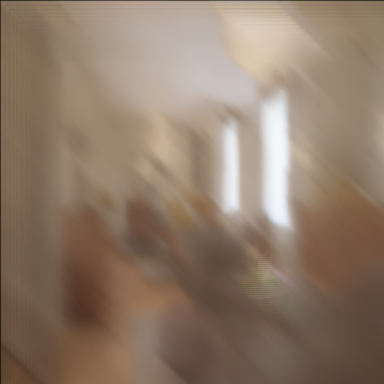} \hfill
    \includegraphics[width=\ftb, height=\ftb,clip,trim={25px 0 0 25px}]{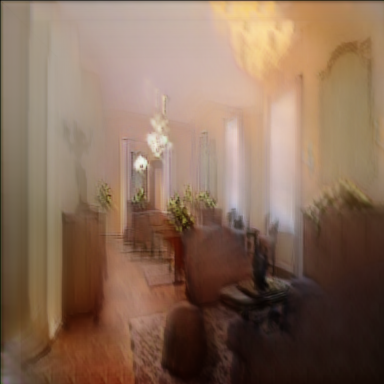} \hfill
    \includegraphics[width=\ftb, height=\ftb,clip,trim={25px 0 0 25px}]{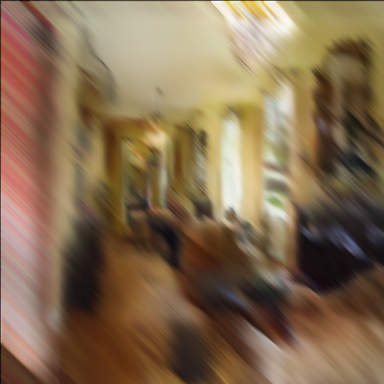} \hfill
    \includegraphics[width=\ftb, height=\ftb,clip,trim={25px 0 0 25px}]{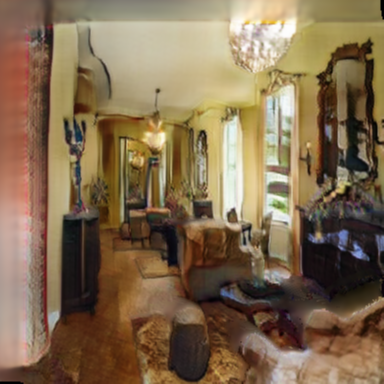} \hfill
    \includegraphics[width=\ftb, height=\ftb,clip,trim={25px 0 0 25px}]{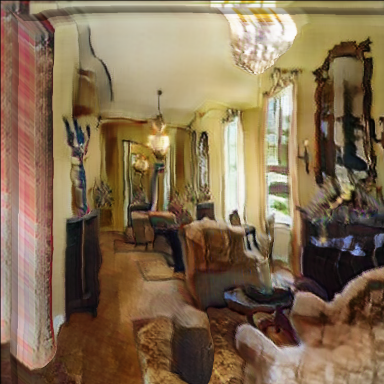}   \\
						   
						   
    \includegraphics[width=\ftb, height=\ftb,clip,trim={25px 0 0 25px}]{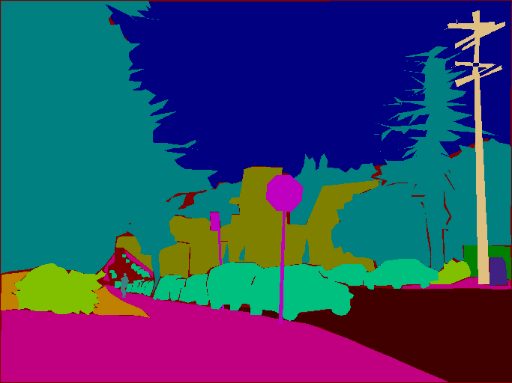} \hfill
    \includegraphics[width=\ftb, height=\ftb,clip,trim={25px 0 0 25px}]{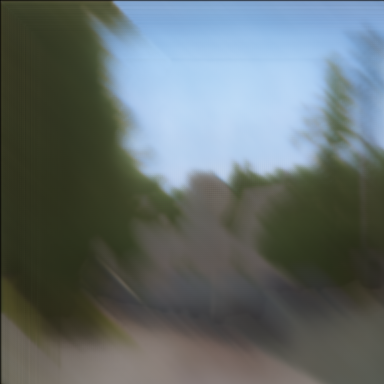} \hfill
    \includegraphics[width=\ftb, height=\ftb,clip,trim={25px 0 0 25px}]{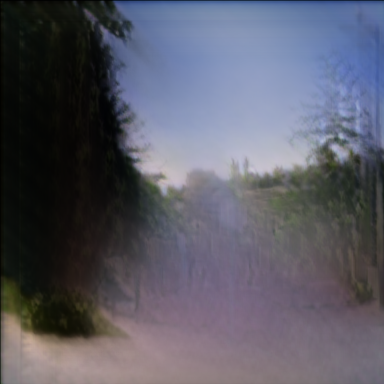} \hfill
    \includegraphics[width=\ftb, height=\ftb,clip,trim={25px 0 0 25px}]{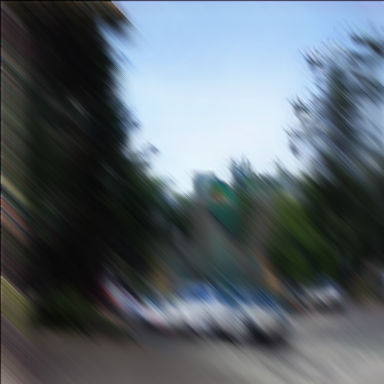} \hfill
    \includegraphics[width=\ftb, height=\ftb,clip,trim={25px 0 0 25px}]{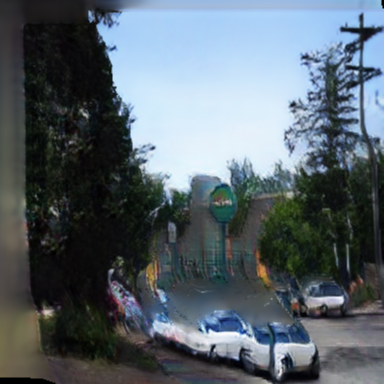} \hfill
    \includegraphics[width=\ftb, height=\ftb,clip,trim={25px 0 0 25px}]{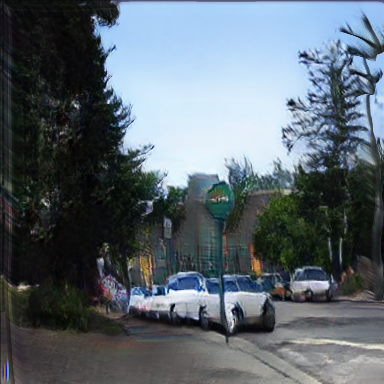}   \\
						   

    \includegraphics[width=\ftb, height=\ftb,clip,trim={25px 0 0 25px}]{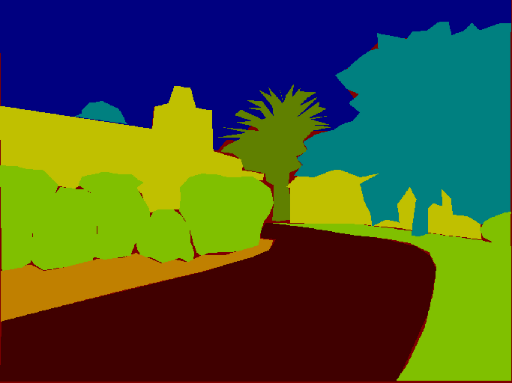} \hfill
    \includegraphics[width=\ftb, height=\ftb,clip,trim={25px 0 0 25px}]{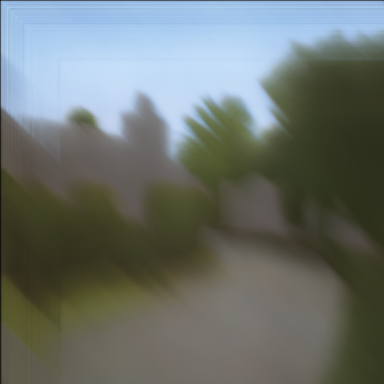} \hfill
    \includegraphics[width=\ftb, height=\ftb,clip,trim={25px 0 0 25px}]{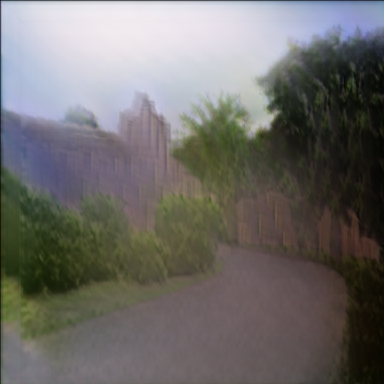} \hfill
    \includegraphics[width=\ftb, height=\ftb,clip,trim={25px 0 0 25px}]{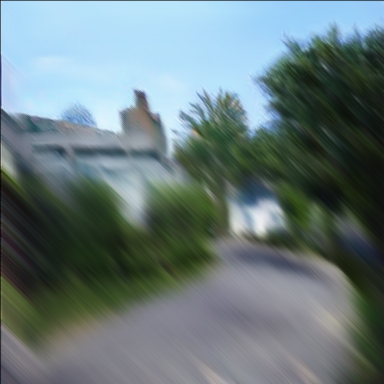} \hfill
    \includegraphics[width=\ftb, height=\ftb,clip,trim={25px 0 0 25px}]{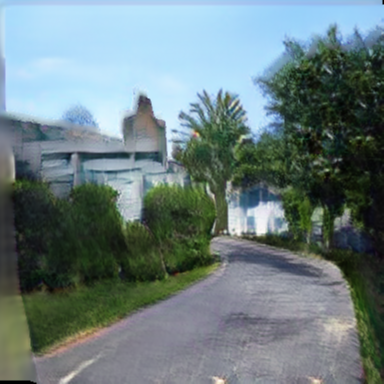} \hfill
    \includegraphics[width=\ftb, height=\ftb,clip,trim={25px 0 0 25px}]{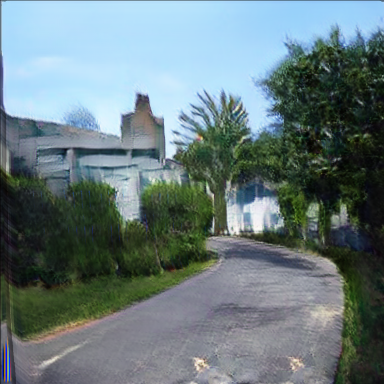}   \\
																									 
																									
%
%
%
%

    \mpage{0.14}{\scriptsize Label} \hfill
    \mpage{0.14}{\scriptsize Direct (U-Net)} \hfill
    \mpage{0.14}{\scriptsize Direct (SPADE)} \hfill
    \mpage{0.14}{\scriptsize Cascade (MPI)} \hfill
    \mpage{0.14}{\scriptsize Cascade (KB)} \hfill
    \mpage{0.14}{\scriptsize Ours}
    \captionof{figure}{\textbf{Visual comparisons.} We compare the generated novel view images of four other baselines and our model among ADE20K and ADE20K-outdoor datasets. The left column shows the input label at the center viewpoint. 
    }
    \label{fig:result}
\end{center}
\end{figure}

%% file: figure/tab_ablation.tex
\begin{table}[t!]
    \centering
    \scriptsize

	\caption{\tb{Ablation study.}
	(a) FID scores under different numbers of depth layers. 
	(b) FID scores of replacing the MPI prediction with per-frame background inpainting. 
	We use NYU dataset for this experiment.
	}
	
\hfill
\mpage{0.45}{(a) Number of depth layers.} \hfill
\mpage{0.5}{(b) Handling dis-occlusion.}
\hfill
\\
\hfill
\mpage{0.47}{
    \begin{tabular}{l|ccc} 
    \toprule
    & \multicolumn{3}{c}{Camera movement} \\
     & (0, 0) & (0.1, 0.1) & (0.2, 0.2)  \\ 
    \midrule
		128 & \textbf{188.83} & \textbf{191.04} & 207.70 \\ 
		64 & 190.70 & 193.71 & 210.06  \\ 
		32 & 190.60 & 194.73 & \textbf{205.59} \\ 
    \bottomrule
    \end{tabular} 
}\hfill
\mpage{0.5}{
\begin{tabular}{l|ccc} 
    \toprule
    & \multicolumn{3}{c}{Camera movement} \\
     & (0, 0) & (0.1, 0.1) & (0.2, 0.2) \\
    \midrule
Ours & \textbf{188.83} & \textbf{191.04} & \textbf{207.70} \\
Diffusion & 190.63 & 192.55 & 210.16 \\
GatedConv & 190.67 & 192.83 & 210.00 \\
    \bottomrule
    \end{tabular}
    }
\hfill

    \label{tab:ablation}
    
\end{table}

%% file: figure/fig_ablation.tex
\begin{figure}[t!]
    \centering
    \mpage{0.14}{\includegraphics[width=\linewidth,clip,trim={256px 0 0 256px}]{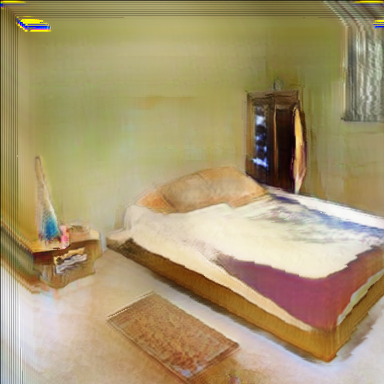}}   
	\mpage{0.14}{\includegraphics[width=\linewidth,clip,trim={256px 0 0 256px}]{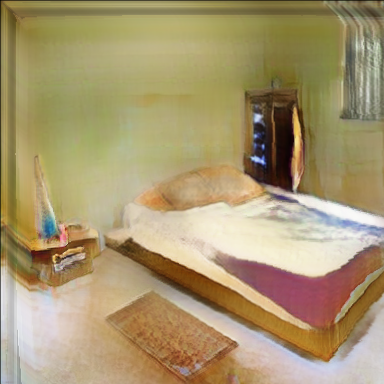}}    
	\mpage{0.14}{\includegraphics[width=\linewidth,clip,trim={256px 0 0 256px}]{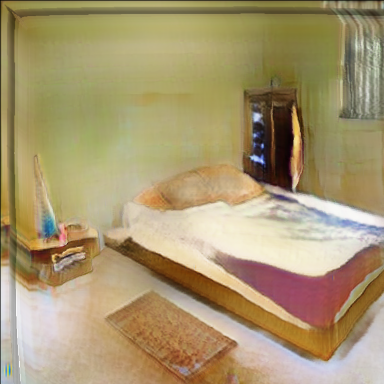}}
    \mpage{0.14}{\includegraphics[width=\linewidth,clip,trim={60px 144px 228px 144px}]{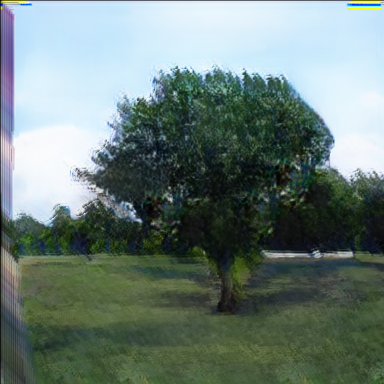}}
    \mpage{0.14}{\includegraphics[width=\linewidth,clip,trim={60px 144px 228px 144px}]{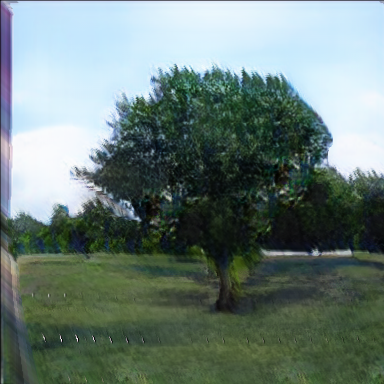}}
    \mpage{0.14}{\includegraphics[width=\linewidth,clip,trim={60px 144px 228px 144px}]{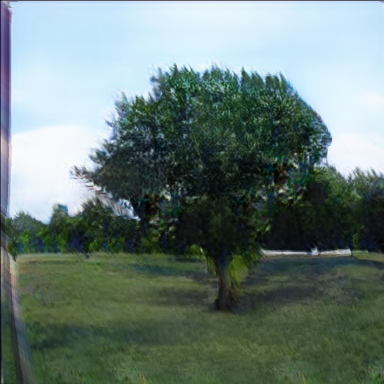}}\\
    
    \mpage{0.14}{\small K=32}
    \mpage{0.14}{\small K=64}
    \mpage{0.14}{\small K=128}
    \mpage{0.14}{\small K=32}
    \mpage{0.14}{\small K=64}
    \mpage{0.14}{\small K=128}

    \caption{\textbf{Number of depth layers.} Increasing the number of depth levels improves the rendered quality.
    }
    \label{fig:ablation}
\end{figure}

%% file: figure/fig_dis.tex
\begin{figure}[t!]
    \centering
    \mpage{0.175}{\includegraphics[width=\linewidth,clip,trim={25px 0 0 25px}]{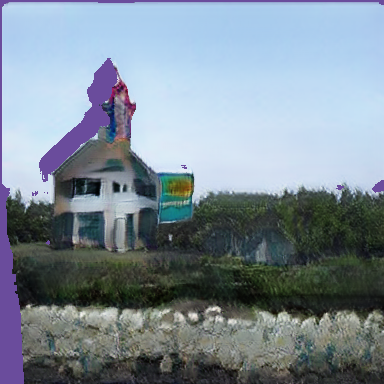}}   
	\mpage{0.175}{\includegraphics[width=\linewidth,clip,trim={25px 0 0 25px}]{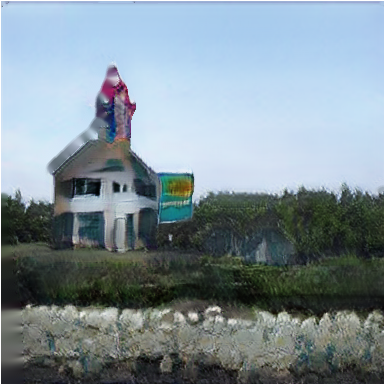}}    
	\mpage{0.175}{\includegraphics[width=\linewidth,clip,trim={25px 0 0 25px}]{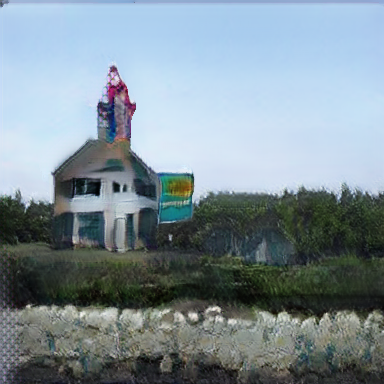}}
    \mpage{0.175}{\includegraphics[width=\linewidth,clip,trim={25px 0 0 25px}]{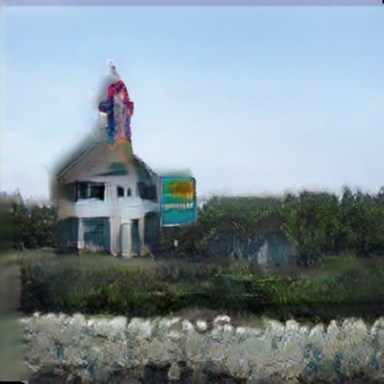}}
    \mpage{0.175}{\includegraphics[width=\linewidth,clip,trim={25px 0 0 25px}]{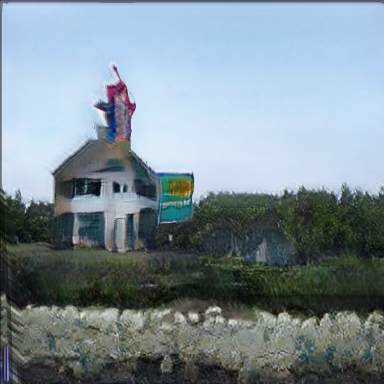}} \\

    \mpage{0.175}{\includegraphics[width=\linewidth,clip,trim={25px 0 0 25px}]{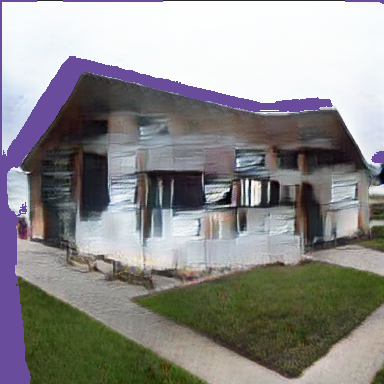}}   
	\mpage{0.175}{\includegraphics[width=\linewidth,clip,trim={25px 0 0 25px}]{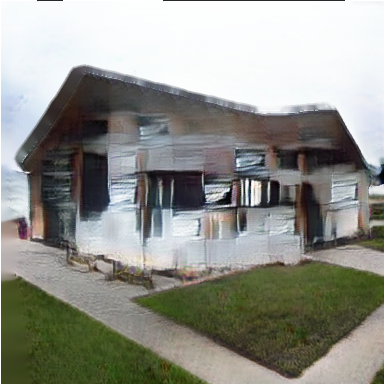}}    
	\mpage{0.175}{\includegraphics[width=\linewidth,clip,trim={25px 0 0 25px}]{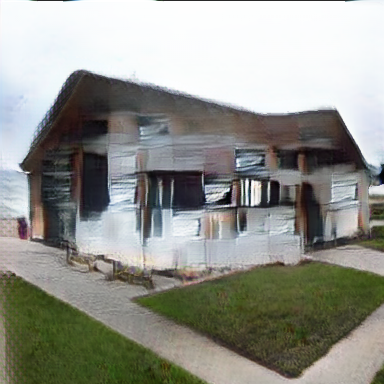}}
    \mpage{0.175}{\includegraphics[width=\linewidth,clip,trim={25px 0 0 25px}]{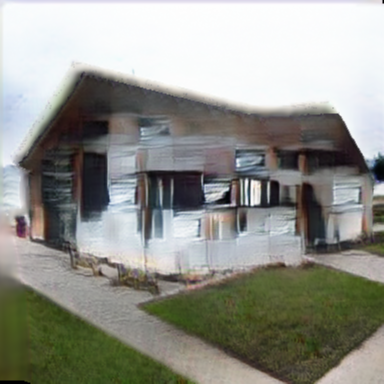}}
    \mpage{0.175}{\includegraphics[width=\linewidth,clip,trim={25px 0 0 25px}]{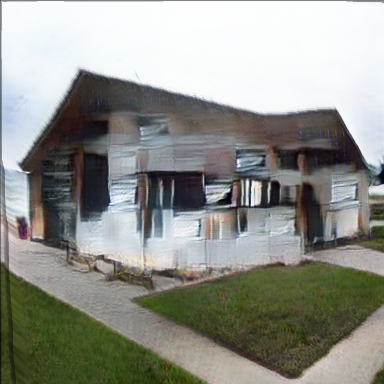}} \\

    \mpage{0.175}{\scriptsize Foreground only}
    \mpage{0.175}{\scriptsize Diffusion}
    \mpage{0.175}{\scriptsize GatedConv \cite{gate}}
    \mpage{0.175}{\scriptsize KB \cite{ken}}
    \mpage{0.175}{\scriptsize Ours}

    \caption{\textbf{Disocclusion handling.} The purple regions (left) are the dis-occluded region. \emph{Diffusion} and \emph{GatedConv} produce artifacts. The 3D Ken Burns method \cite{ken} generates blurry and unnatural dis-occluded contents. Our model hallucinates visually appealing results.
    }
    \label{fig:dis}
\end{figure}

%% file: figure/fig_user.tex
\begin{wrapfigure}{R}{0.4\linewidth}
\begin{center}
\includegraphics[width=1.0\linewidth]{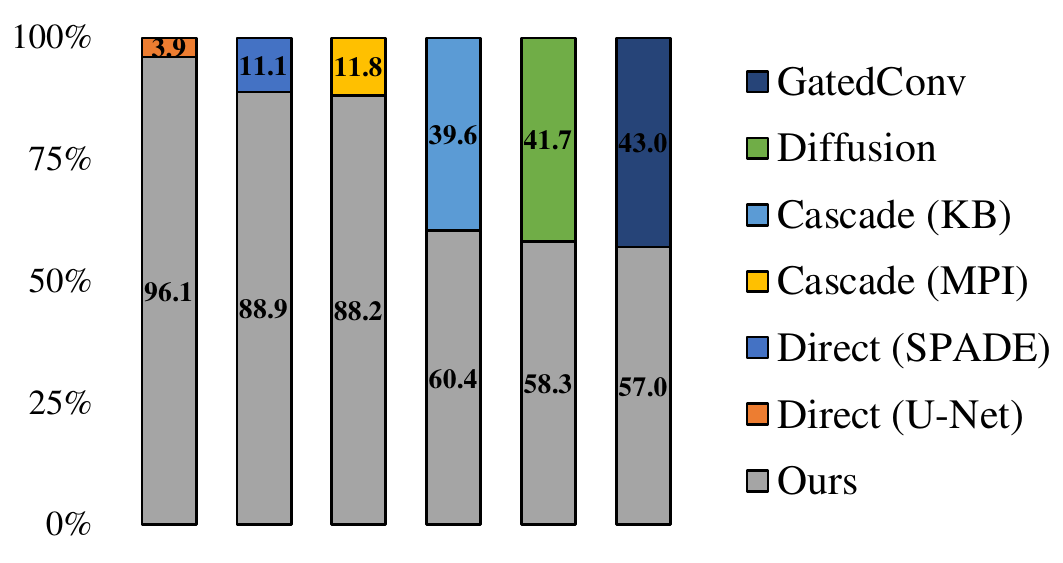} 
\end{center}
\caption{
\textbf{User study.} We show the user preference between the proposed method and baselines.
}
\label{fig:user}
\end{wrapfigure}

%% file: 5_conclusion.tex
\section{Conclusions}
\label{sec:conclusions}

We have introduced a new problem called semantic view synthesis.
The problem aims to generate a photorealistic image from a given semantic label map that supports novel view rendering. 
The new form of visual content creation offers significantly more immersive experience than the conventional 2D image synthesis task.
This is technically achieved by carefully integrating techniques from semantic image synthesis and view synthesis.
Our core idea is to model the 3D scene by first modeling the visible surface then further inferring the full 3D scene representation.
We conduct an extensive experimental evaluation to validate our model design and show favorable results over several baseline methods.

%% file: rebut_content.tex

# AnonReviewer3

**Performance comparison with Diffusion and GatedConv**

The diffusion-based and GatedConv-based in-painting approaches do produce slightly more realistic images *for single views* in Table 1(b).
However, as the images at different views are processed independently, these methods do not consider the consistency across different views. 
As demonstrated in the supplementary video (see background inpainting section), these methods produce temporal flickering artifacts due the the lack of cross-view consistency.
**User study**

Thank you for your suggestion. 
We conducted a perceptual user study to quantify the user preference over the proposed method and the six baseline approaches.
For each test during the study, we present two novel view videos of the same scene generated by two different methods with circular camera motion (in randomized order). 
We then ask the participant to select his/her preferred result. 
Each participants complete $60$ pairwise comparisons. 
\jiabin{How many videos were used? Which dataset?}
In total, we conduct the study with $47$ participants ($2820$ binary votes).
The results shown below validate that the proposed method synthesizes more realistic novel view videos compared to the baseline approaches.
We will include the results in the revised paper.

|Comparison|User preference (
|-|-|
|Ours vs. Direct (U-Net)|96.1 vs.  3.9|
|Ours vs. Direct (SPADE)|88.9 vs. 11.1|
|Ours vs. Cascade (MPI) |88.2 vs. 11.8|
|Ours vs. Cascade (KB)  |60.4 vs. 39.6|
|Ours vs. Diffusion     |58.3 vs. 41.7|
|Ours vs. GatedConv     |57.0 vs. 43.0|


**ADE20K indoor split**

We follow the evaluation protocol of SPADE [29] to conduct the experiments on the ADE20K and ADE20K-outdoor [31] datasets.
We agree that an indoor/output split would make more sense. 
However, the outdoor split from [31] does not provide the corresponding data ID in the original ADE20K test set.
As a result, we cannot obtain the indoor split by removing the outdoor split from the original ADE20K dataset.

#AnonReviewer2


**Additional experiments using inferred depth map for predicting background layer and the blending weights**

Thank you for the suggestion. 
Originally, we thought that using the inferred depth for computing the alpha map is sufficient for guiding the MPI generator $G_m$ for predicting background color and blending weights.
As suggested, we concatenate the inferred depth map $d$ with the inferred color image $x^s$ as input for the MPI generator. 
Below we show the results of this model (2nd row) and compare it with our previous model along with two other baselines in Table 1(b) in the paper. 
The results shows that leveraging depth cues in the MPI generator leads to moderate improvement over our previous model (with only color image as the input). 
We will include the complete results in the revised manuscript.

|-|(0,0)|(0.1,0.1)|(0.2,0.2)|(0.3,0.3)|
|-|-|-|-|-|
|Ours (color only)      | 121.23| 122.83| 132.43| 153.36|
|Ours (color and depth) | 120.30| 122.06| 131.32| 138.44|
|Diffusion | 121.94| 123.82| 129.15| 138.71|
|GatedConv | 121.94| 123.82| 129.17| 138.79|

**Applications**

Semantic view synthesis enables non-expert users to imagine and create photorealistic visual contents with free-viewpoint rendering from simple semantic layout.
Creating such contents may require a high level of artistic skills and efforts. 
While the visual quality of the synthesized images do not reach the level of realism for practical applications yet, it is promising given the rapid progress on generative models in recent years.
We hope our work will inspire future research and allow everyone to express their imagination.



**Additional experiments with Savva et al. ICCV 2019**

Thanks for the pointer. 
We will modify the description in Line 213-215 and cite this paper. 
As suggested, we will also test our model on this dataset and try using the semantic labels and depth pairs to train our model.



**Additional comparison with Tucker and Snavely: Single-view view synthesis with multiplane images, CVPR 2020**

We will ask the authors for a reference implementation and include the comparison in the revised paper.

**Capitalization in the bibliography**

Thank you! We will fix the title issues in the bibliography.


#AnonReviewer1

**Details about depth**

We train our disparity generator to predict the absolute disparity values, as described in Line 72-84 and Line 106-110 in the supplementary document. 
Specifically, we first use the pre-trained MiDaS model [22] to obtain *relative* disparity maps. 
We then use the prediction from the DPSNet [17] model to transform the predicted relative disparity images to absolute disparity images. 
The absolute disparity images after the transformation serve as the pseudo ground-truth images for training our disparity generator.

**Additional experiments with GAN loss**

Thanks for the suggestions.
We add the GAN training loss and find that it indeed improves the image quality. 
Here we show the results of adding GAN loss with a weight of 0.01 (2nd row).
We will include the complete results in the revised paper.

|-|(0,0)|(0.1,0.1)|(0.2,0.2)|(0.3,0.3)|
|-|-|-|-|-|
|Ours     | 121.23| 122.83| 132.43| 153.36|
|Ours + GAN loss | 121.25| 121.74| 130.12| 137.25|
|Diffusion| 121.94| 123.82| 129.15| 138.71|
|GatedConv| 121.94| 123.82| 129.17| 138.79|


**Additional comparison with Wiles et al: End-to-end View Synthesis from a Single Image, CVPR 2020**

The official source code (released on Github at 3/25) was not available before the ECCV submission deadline. 
We will include a comparison to this approach in the revised paper.

**Proofreading**

Thank you! We will correct the typos in the revised draft.
